\documentclass[a4paper,twocolumn]{article}

\usepackage[latin1]{inputenc}
\usepackage{float}
\usepackage{natbib}
\usepackage{times}

\newcommand{\binom}[2]{C_{#1}^{#2}}

\usepackage{graphicx}

\begin{document}
\bibliographystyle{plainnat}
\title{(The Final) Countdown\\{\em Preprint}}
\author{Jean-Marc Alliot\footnote{Institut de Recherche en
    Informatique de Toulouse}}
\date{}
\maketitle

\begin{abstract}
  The Countdown game is one of the oldest TV show running in the
  world. It started broadcasting in 1972 on the french television
  and in 1982 on British channel 4, and it has been running since in
  both countries.
  The game, while extremely popular, never received any serious
  scientific attention, probably because it seems too simple at first sight.
  We present in this article an in-depth analysis of the
  numbers round of the countdown game. This includes a complexity
  analysis of the game, an analysis of existing algorithms, the presentation
  of a new algorithm that increases resolution speed by a factor of
  20. It also includes some leads on how to turn the game into a
  more difficult one, both for a human player and for a computer, and
  even to transform it into a probably undecidable problem.
\end{abstract}

\section{Introduction}
The Countdown~(\cite{Countdown}) game is one of the oldest TV show running in the
world. It started broadcasting in 1972 on the french television as
{\em ``des chiffres et des lettres''}, literally {\em ``numbers and
  letters''} with a numbers round called
{\em ``Le compte est bon''}, literally {\em ``the count is good''}). It
started broadcasting
in 1982 on British channel 4 as {\em ``Countdown''}, and it has been
running since in both countries.

The numbers round of the game is extremely simple: 6 numbers are drawn
from a set of 24 
which contains all numbers from 1 to 10 (small numbers) twice plus 25, 50, 75 and
100 (large numbers). Then, with these six numbers, the contestants have to find a
number randomly drawn between 101 and 999\footnote{Whether 100 is a
  possible number to search is a matter of controversy. It seems like
  it could be in the UK, but not in France, so we decided to let it
  out.}, or, if it is impossible, 
the closest number to the number drawn. Only the four standard
operations ($+\,-\,\times\,/$) can be used. As soon as two numbers have been 
used to make a new one, they can't be used again, but the new number
found can be used. For example, if the six numbers 
drawn are {\em 1,1,4,5,6,7} and the number to find is {\em 899} the
answer is:
{\small
\begin{verbatim}
  Operations   Remaining
 6 x  5 = 30  {1,1,4,7,30}
30 +  1 = 31  {1,4,7,31}
 4 x  7 = 28  {1,28,31}
28 +  1 = 29  {29,31}
29 * 31 = 899 {899}
\end{verbatim}
}
There are usually different ways to find a
solution. 
The {\em simplest answer} is usually defined as the answer using the
least number of operations, and if two solutions have the same number
of operations, a possible refinement is to keep the one having the
smallest highest number\footnote{There are a few differences
  between the french and the British game. In the french version, all
  numbers are drawn at random while in the British game, the
  contestants can choose how many large numbers will be present in the
six numbers set.}. 

The game, while extremely popular, never
received any serious scientific attention. There was a very early
article in the french magazine ``l'Ordinateur Individuel'' in the late
seventies, written by Jean-Christophe Buisson~(\cite{Buisson}), which
described a simple algorithm. The only article written on
the subject in English was published twice~(\cite{Defays90,Defays95}) by
Daniel Defays. Defays also published in 1995 a book in
french~(\cite{Defays95f}) which used the game as a central example for
introducing artificial intelligence methods. But the ultimate goal of
Defays was not to develop an accurate solver for the game, but a
solver mimicking human reasoning (such as the {\em Jumbo}
program by Hofstadter), including possible mistakes (in French, Defays
sometimes named his program {\em ``le compte est mauvais''},
literally {\em the count is bad}, a joke on the original name of the
game, indicating that it might make mistakes while searching for the
solution).

There are many commercial or free programs developed for this
game. Some of them are bugged or use incomplete or incorrect
algorithms. Many 
websites in France and in Great Britain discuss the game and how to
program it, with lot of code, lot of statistics, and sometimes lot of
errors. The first goal of this article is to do a scientific analysis of the
game regarding its complexity and to provide a set of cutting edge
algorithms and codes to solve it properly. Its second goal is to
investigate potential extension of the games, either to turn it into a
more complex problem, or into a (maybe) undecidable problem
on some of its instances.

We use a few mathematical symbols and functions in this paper: $n!$ is
the factorial of n, $\binom{n}{p}=\frac{n!}{p!\,(n-p)!}$ is the number
of subsets having $p$ elements in a set of $n$ distinct elements,
$\Gamma(z)$ is the Euler Gamma function, $E(x)$ is the integer part of $x$.

\section{Elementary algorithms\label{seccomp}}
\subsection{Decomposition in sub-problems}
The first published algorithm~(\cite{Buisson}) used a simple
decomposition mechanism. Let's consider the following example: numbers 
{\em 3, 50, 7, 4, 75, 8}, number to find {\em 822}. The algorithm
would start from the solution (822) and use a backward chaining approach in the
Prolog way. However, not all operations were tried; at odd steps, only
addition and subtractions were used, while at even steps only
divisions were used. So the algorithm would at the first step 
generate thirteen numbers: $822$, $822 \pm 3$, $822\pm 50$, \ldots, and then
try to divide all of them by the remaining 5 (or 6 if no number was
added or subtracted) numbers. If a division succeeds, the algorithm
would then be applied recursively on the new result with the remaining
numbers. Here the solution can be found by:
{\small
\begin{verbatim}
   Operations         Remaining
(822 + 50) / 4 = 218  {3,7,75,8}
(218 +  7) / 3 =  75  {75,8}
75 - 75 = 0           {8}
\end{verbatim}
}
When 0 is reached the solution has been found.

The complexity of this algorithm is very low. If we have $n$ numbers,
we first generate $2n+1$ numbers and try to divide them by $n-1$
numbers, so we have to do $(2n+1)(n-1)$ trial divisions.
At the next step, we would have on the
average $2(n-2)+1$ numbers to divide by $n-3$ numbers, so
$(2(n-2)+1)(n-3)$ trial divisions, and so on. 

On the one hand, if all divisions succeed, we have a maximal
complexity of:
\begin{eqnarray*}
\prod _{i=1}^{n/2} (4 i+1) (2 i-1)&=&\frac{2^{\frac{3 n}{2}} \Gamma
  \left(\frac{n}{2}+\frac{5}{4}\right) \Gamma
  \left(\frac{n+1}{2}\right)}{\sqrt{\pi } \Gamma
  \left(\frac{5}{4}\right)}\\
&\simeq& 8^{\frac{n}{2}}\,((\frac{n}{2})!)^2
\end{eqnarray*}

On the other hand, if only one division succeeds at each step, the
minimal complexity is:
$\sum_{i=1}^{n/2} (4 i+1) (2 i-1)=\frac{n^3+9n-4}{12}$
For $n=6$, we have a maximal complexity of 8775
trial divisions, and a minimal complexity of 97 trial divisions. 

This algorithm was popular in the seventies, when machines were slow,
with only 8 bit addition and subtraction in hardware,
with  division and multiplication implanted in software on
microprocessors.
Moreover, programs were often
written using slow, interpreted languages. Some of the initial
programs were written in Basic or in Prolog, and could not handle a
large number of computations in the 30 or 45s allowed by the game.
Indeed, this algorithm was published again in
1984~(\cite{Froissart84}) in another journal, which means that even 4
years later, few people were able to write programs to solve
completely the game.

This
algorithm has of course serious drawbacks. It is
impossible to compute solutions requiring intermediate results, such
as the first one presented in this article, because 31 and 29 must be
built independently before multiplying them to have 899. It is even
impossible to find solutions with divisions. Moreover,
this method can only find the exact result. If it doesn't exist, the
computation has to be restarted with the closest number to the number
to find as a new goal.

This algorithm was later refined with faster machines by using all
possible operations at each step.
At
the first step of the algorithm there are 6 numbers available and 4
possible operations, which would give 24 numbers at most (here:
$822+3$, $822-3$,
$822\times 3$, $822/3$ and so on with 50, 7, 4\ldots). Division is
not always possible, and  so there are in fact between 18 and 24
numbers (here there are only 19 numbers at the first step, as 822 can
only be divided by 
3). This algorithm is recursively applied until 0 is found or until no
number remains in the pool of available numbers.

Here, the solution can be found in the following way:
{\small
\begin{verbatim}
822 + 50 = 872 {3,7,4,75,8}
872 /  4 = 218 {3,7,75,8}
218 +  7 = 225 {3,75,8}
225 /  3 =  75 {75,8}
 75 - 75 =   0 {8}
\end{verbatim}
}

The maximal complexity of the algorithm is $(6\times 4)\times (5\times
4)\times\cdots(1\times 4) = 6! \, 4^{6}$ If we consider the general
case with $n$ numbers the complexity is $n! \, 4^{n}$. For $n=6$, the
maximal number of operations is $491520$. If we consider that the
actual number of operations at each step is closer to 3 than to 4, we
have a minimal complexity of $n! \, 3^{n}$, and for $n=6$ the minimal
number of operations is $87480$. Let's also keep in mind that
even if division is not possible it has to be tested before being
discarded, so this minimal complexity is an inferior bound that can
never be reached.

This refinement adds more solutions but it is still impossible to find
solutions requiring intermediate results, and impossible to find
directly approximate results.

\subsection{The depth first algorithm\label{secdepth}}
The recursive depth first algorithm is extremely easy to
understand. Let's consider the complete set of $n$ numbers. We simply pick
two of them 
($\binom{n}{2}=\frac{n(n-1)}{2}$ 
possibilities) and
combine them using one of the four possible 
operations. The order of the two numbers picked is irrelevant as the order does
not matter for the addition and the multiplication ($a+b=b+a$ and
$a\times b=b\times a$), and we can only use one order for the other
two operations (if $a > b$,
we can only compute $a-b$ and $a/b$, and if $a < b$, $b-a$ and
$b/a$). Then we put back the result of the computation in the set, 
giving a new set of $n-1$ numbers. We just repeat the algorithm until no
number remain in the pool and then backtrack to the previous point of
choice, be it a number or an operation. This is a simple
depth-first search algorithm, which is exhaustive as it searches the
whole computation tree.

The maximal complexity of the algorithm is given by:
$ (\frac{n\times(n-1)}{2}\times 4) \times
(\frac{(n-1)\times (n-2)}{2}\times 4) \times \cdots\times
(\frac{2\times 1}{2}\times 4)$. This gives:
\begin{eqnarray}
  \label{dmax}
  d_{max}(n)&=&n!\,(n-1)!\,2^{n-1}\\
  \label{dmin}
  d_{min}(n)&=&n!\,(n-1)!\,(\frac{3}{2})^{n-1}
\end{eqnarray}
For $n=6$, we have a maximal number of $2764800$
operations and a minimal number of $656100$ operations.
The algorithm is extremely easy to implement in this naive
version. No complex data structures are needed, and being a depth
first algorithm, it requires almost no memory.

The first recorded implementation of this algorithm~(\cite{Alliot86})
was developed for an Amiga 1000 (a MC68000 based microcomputer with a 7MHz
clock). It was written in assembly language and solved the
game in less than 30s. However, this implementation was not perfect,
as it worked only with unsigned short integers (integers between 0 and
65535), and was thus unable to compute numbers that required
intermediate results higher than 65535 (and there are some, such as
finding 996 with $\{3,3,25,50,75,100\}$ which requires using 99600 as an
intermediate result).

\subsection{The breadth first algorithm\label{breadcomp}}
The breadth first algorithm is a little bit more difficult to
understand. It is also a recursive algorithm, but it works on the
partitions of the set of numbers. The first presentation of this
algorithm seems to be \cite{Pin98}.
\begin{itemize}
\item First, we create all sets generated by only one element.
  With the same example, we have of course 6 elements $g(\{3\})=\{3\}$,
  $g(\{50\})=\{50\}$, 
  $g(\{7\})=\{7\}$, $g(\{4\})=\{4\}$, $g(\{75\})=\{75\}$, $g(\{8\})=\{8\}$
\item Next we create the sets of all numbers that can be computed
  using only two numbers. Here for example all the numbers generated
  by $\{3,50\}$ are the elements of  $g(\{3\})$ applied to the
  elements of
  $g(\{50\})$ which give the set $g(\{3,50\})=g(\{3\}) . g(\{50\}) =
  \{53,47,150\}$. $\{3\}$ and $\{7\}$ give 
  $g(\{3,7\})=\{10,4,21\}$. $\{50\}$ and $\{7\}$ give
  $g(\{50,7\})=\{57,43,350\}$. We will have $\binom{6}{2}$ such sets.
\item Next we create the sets of all numbers that can be computed
  using only 3 numbers. For example, the set of numbers generated by 
  the 3 numbers 3, 50 and 7 is
  $g(\{3,50,7\}) = g(\{3\}) . g(\{50,7\}) \cup g(\{50\}) . g(\{3,7\})
  \cup g(\{7\}) . g(\{50,3\})$ Here for example
  $g(\{3\}) . g(\{50,7\}) = \{54,60,171,19,40,46,129,347,353,1050\}$ 
  There are
  $\binom{6}{3}$ such sets.
\item The algorithm proceeds with all sets generated by 4 numbers. For
  example the set generated by $\{3\}$, $\{50\}$, $\{7\}$ and $\{4\}$
  is $g(\{3,50,7,4\}) = g(\{3\}) . g(\{50,7,4\}) \cup g(\{50\}) . g(\{3,7,4\}) \cup
  g(\{7\}) . g(\{50,3,4\}) \cup g(\{4\}) . g(\{3,50,7\}) \cup
  g(\{3,50\}) . g(\{7,4\}) \cup 
  g(\{3,7\}) . g(\{50,4\}) \cup g(\{3,4\}) . g(\{50,7\})$ There are
  $\binom{6}{4}$ such sets. 
\item We proceed with all sets generated by 5 numbers, applying
  exactly the same algorithms. There are $\binom{6}{5}$ such sets.
\item Then we create the set generated by all six numbers.
\end{itemize}

The complexity of this algorithm is not so easy to compute. It is 
sometimes mistakenly presented as being $2^n$~(\cite{Mochel03}), but it is a
very crude estimation.

If we call
$N(p)$ the number\footnote{The $N(p)$ are probably related to Bell
  numbers, but they are not the same} of elements in a set generated
by $p$ elements, the
total number of operations will be $\sum_{p=1}^n \binom{n}{p}
N(p)$. We still have to compute $N(p)$. It is possible to establish a
recurrence relationship between $N(p)$ and $N(p-1)$, $N(p-2)$,
etc. Let's see that on 
an example. $N(4)$ is the sum of two terms:
\begin{itemize} 
\item $N(3)\times N(1)\times 4\times \binom{4}{3}$ which is
the number of elements in a set built by combining with the 4
operations a set having $N(1)$ elements and a set having $N(3)$
elements. There are $\binom{4}{3}=4$ such numbers. For example for
$\{1,2,3,4\}$, we have  $\{1,2,3\} . \{4\}$, $\{1,2,4\} . \{3\}$,
    $\{1,3,4\} . \{2\}$, $\{2,3,4\} . \{1\}$
\item $N(2)\times N(2)\times 4\times (\binom{4}{2}/2)$ For example,
  for $\{1,2,3,4\}$ we combine $\{1,2\} . \{3,4\}$, $\{1,3\}
  . \{2,4\}$ and $\{1,4\} . \{2,3\}$
\end{itemize}
More generally, we have:

$N(p) = (\sum_{i=1}^{p-1} \binom{p}{i} N(i) N(p-i))/2\times4$ 

A simple computation gives: 

$N(p)=4^{p-1} \prod_{i=1}^{p-1} (2i-1)$

And thus the complexity for $n$ numbers is:
\begin{eqnarray}
  \label{bmax}
  b_{max}(n)&=& \sum_{p=1}^n \binom{n}{p} \,4^{p-1} \prod_{i=1}^{p-1} (2i-1)\\
  \label{bmin}
  b_{min}(n)&=& \sum_{p=1}^n \binom{n}{p} \,3^{p-1} \prod_{i=1}^{p-1} (2i-1)
\end{eqnarray}
For $n=6$ we have a maximal number of 1144386 operations, half the
number of the operations required by the depth first algorithm, and a
minimal number of 287331 operations.
  
%

\section{Implementation and refinements\label{implem}}
To compare the algorithms, the programs were all written in Ocaml~(\cite{Ocaml}).
The implementation was not parallel and the programs were run on a
980X. For very large instances, an implementation of the best
algorithm (depth first with hash tables) was written in C and assembly
language.
MPI~(\cite{MPI}) was used to solve problems in parallel and the
program was used on a 640 AMD-HE6262 cores cluster using 512 cores.
With the same algorithm, the C program on a
single core is twice faster than the Ocaml program.

The 980X used in this section is a 6 cores Intel processor running at
3.33Ghz (a clock 
cycle of 0.3ns) with a 32kb+32kb L1 cache by
core, a 256kb L2 cache by core and 12Mb of 
L3 (Last Level Cache or LLC) cache common to all
cores. Memory timings~(\cite{Levinthal09}) for the Core i7 family
and Xeon 5500 family are roughly of 4 clock cycles for L1 cache and 10 cycles for
L2 cache. L3 cache access times depend on
whether the data is local to the core (40 cycles), shared with another core
(65 cycles) or modified by another core (75 cycles). Here, the
application is completely local to one core, so it is safe to assume
an access time of 40 cycles. Outside the L3 cache, access times
depend on the number and type of DIMMs, frequency of the memory bus,
etc\ldots A good guess is around 60ns, which is around 5 times slower
than the L3 cache (40 cycles takes approximately 12ns).

In this section we study the standard countdown game: $n=6$ numbers are
drawn from a pool of 24, with all
numbers in the range 1-10 present twice, plus one 25,
one 50, one 75 and one 100. The number of different possible instances is:
\begin{eqnarray*}
&\binom{14}{6}&\mbox{with no pair}\\
+&\binom{10}{1}\times\binom{13}{4}&\mbox{with one pair}\\
+&\binom{10}{2}\times\binom{12}{2}&\mbox{with two pairs}\\
+&\binom{10}{3}&\mbox{with three pairs}\\
=&13243&\\
\end{eqnarray*}
Programs are so fast that trying to accurately measure the execution
time of a single instance is impossible. So, in the rest of this
section, all programs solve the complete set of
instances and the time recorded is the time to complete the entire
set: when a time of 160s is given, the mean time of resolution of one
instance is $160/13243=0.01s$

\subsection{The depth first algorithm}
Implementing the naive version of the depth-first algorithm is a
straightforward process. First the algorithm searches the entire space
with the pool of initial numbers 
and marks all numbers reached as being solvable. Then
if the number to find is marked, it is solvable.
In case of failure, there is no need to start a new search: finding
the closest number marked as solvable in the array is enough.

Storing solutions is easy: each time we compute a number, we first
check if it 
has been found already. If it hasn't been found, we store
the list of operations which led to it (the list to store is just the
branch currently 
searched). If the number has already been found, we store the new
solution if and only if the list of operations for this solution is
smaller than the one already stored. The algorithm being a depth first
algorithm, we have no guarantee that the first solution found is the
shortest.

There are a few simple improvements to implement,
that will be used by all subsequent programs:
\begin{itemize}
\item never divide by 1, because it doesn't generate any new number.
\item never multiply by 1 for the same reason.
\item never subtract two equal numbers
\item if $a-b$ returns $b$ then discard this branch
\item if $a/b$ returns $b$ discard also this branch
\item as stated in the previous section, it is useless to test the
  pair $(a,b)$ and the 
  pair $(b,a)$. If $a>=b$ we just compute $a+b$, $a*b$, $a-b$ and $a/b$
  (if $b$ divides $a$). If $b>a$, we use $b-a$ instead of $a-b$ and $b/a$
  instead of $a/b$.
\end{itemize}

The performance of the algorithm is good: {\bf the native code version
solves the complete problem (the 13243 instances) in 160s}, or a mean
of 0.01s by instance. 

\subsection{The depth first algorithm with hash tables}
The idea is to use for this problem an (old)~(\cite{Zobrist70})
improvement which has been often used in many classical board games:
hash tables. 

First, we notice that, when solving the game, if the same set of
numbers appears a second time in the resolution tree, the branch can be
discarded: as it is a depth first search where the size of the set of numbers
strictly decreases by one at each level in the tree, we know that this
branch has
already been fully developed somewhere else in the tree and that all
possible results have already been computed and all 
numbers that can be found with that set of numbers have already been
marked. We just need a way to uniquely identify an identical set of
numbers, and to do this in a very short amount of time as this test
will take place each time a new number is generated.

There is, as usual with hash tables, a trade-off between generality
(being able to identify any set and store all hash values) and speed
(losing
some generality to be faster). There are two main problems when using
hash tables: computing the hash value and storing/retrieving it.

For computing the hash value, Ocaml provides a generic {\em hash}
function that operates on any
object and returns a positive integer that can be used as an
identifier of the object. However, using this function on sets of
numbers proved to be much too slow. Thus a faster, incremental
approach, was used: an array $h(x)$ of 64 bits random values is
created at the start of the program. Each time a number $x$ is added to
the pool of numbers, $h(x)$ is added to the hash value, and when $x$
is removed from the pool, $h(x)$ is subtracted from the hash
value. This is slightly different from the hash computation  in
board games where the function used is the faster $xor$ function, both
for adding and removing objects. However, $xor$ can not be used here
as two identical numbers can be in the pool at the same time and would
cancel out each other: the set $\{1,1,2,3,4\}$ would have the same
hash value as $\{2,3,4\}$.

Storing the hash value required some experimental tests. Using a
single set structure to store all values was ruled out from the start,
as it would have been way too slow (the $\log n$ access time is
too important). Thus a more classical array structure was chosen,
where a mask of $n$-bits was applied to the 64-bits hash value,
returning an index for this array (the size of the array is of course
$2^n$). 
There were two remaining problems to solve: how to 
handle hash collisions and how large the array must be. 

Hash collisions happen when two different objects having different
hash values have the same hash index. They can be solved in two
(main) ways: maintaining a set of values for each array element, or
having a larger array to minimize hash
collisions. However having a too large array can also have detrimental
consequences: as the access to the hash array is mostly
random, cache
faults are very likely to happen at each access if the array doesn't
fit in the cache. The
largest the part of the array out of the cache, the higher the
probability to have a cache fault and to seriously
slow down the program.

On this processor, the maximal size of an
array of 64 bits integers that would fit in the L2 cache is
$2^{15}=32768$ elements and the maximum size of a 64 bits integer
array that would fit in the L3 cache is $2^{20}=1048576$ elements. It
is important to remember that the L3 cache is shared by all cores, and thus
degradation might (or might not) appear for smaller values as other
processes are running.
\begin{figure}[ht!]
\begin{center}
  \includegraphics[width=7.5cm]{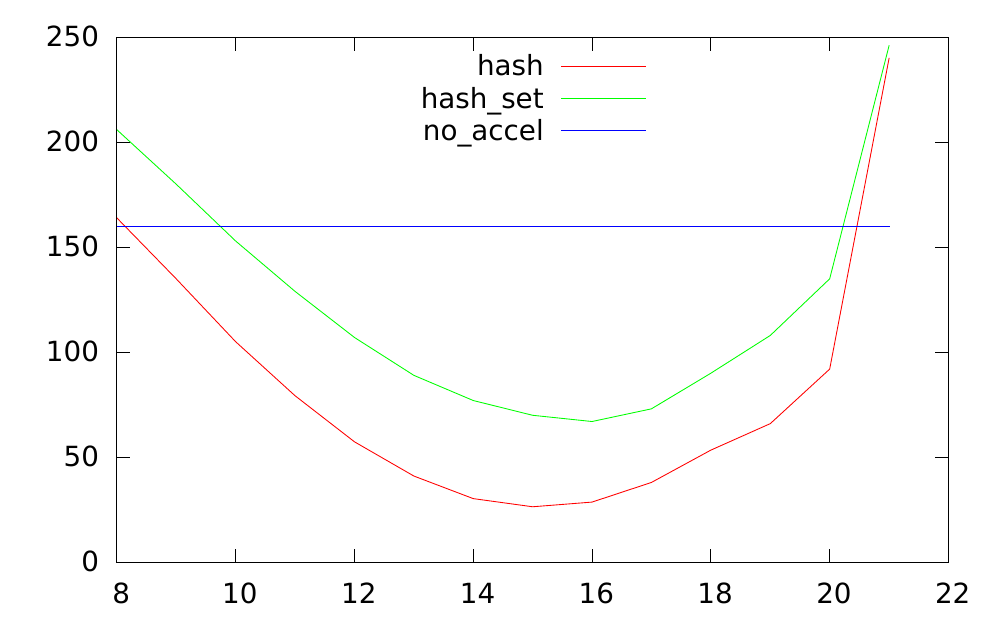}
  \caption{Influence of the size of the hash table\label{fig2}}
\end{center}
\end{figure}
On figure~\ref{fig2}, we have the result of the experimentation. The
x-axis is the size of the hash table in bits, the 
y-axis the time needed to solve the 13243 instances. The blue plot is
the time without hash tables, the 
red one the time with a simple array hash table and the green one the
time with an array containing sets to hold all numbers. 

As expected,
the cache issue is a fundamental one, and results are in
accordance with the theory. Let's concentrate on the red plot,
which is the easiest to interpret. As long as we remain in the L2 cache
(up to 15), increasing the size of the hash table enables to store
more elements and thus to cut more branches in the tree. {\bf For $n=15$,
the 13243 instances are solved in 26s, 5 times faster than without
hash tables}. Over $n=15$, part of the hash table is in the L3 cache and thus,
while we are still cutting more branches as we are storing more
elements, L2 cache faults are slowing the program faster than we are
accelerating it by cutting more branches. For $n=20$, we begin to have
problems to keep the hash table in the L3 cache, and for $n=21$ there
are so many L3 cache faults that the program is slower than what it was
without hash tables. 

The green plot shows that when we store all
results in an array of sets, there are quickly too many elements, and
thus we are never able to remain inside the L2 cache. The minimal time is 67
seconds which is the time we also have for $n=19$ with the simple array
structure when we still fit inside the L3 cache. There are however not too
many elements as data clearly remain inside the L3 cache as long as the size of
the hash array itself is less than the size of the L3 cache. As 
soon as we are out of the L3 cache, the two methods give the same
(bad) results: times are equal for $n=21$ as most of the time is spent
in cache faults.  

This might seem like a strange result but the reason is easy to
understand: the program is doing very little work between two accesses
to the hash table: one arithmetic operation and a few tests, reads and
stores. All these operations use data and code that remain in the L1
cache, and they are thus extremely fast. Then, memory accesses can
become the bottleneck of the program.
Let's also remember that the number of
generated positions with the depth-first algorithm is between 
$6!\,5!\,(\frac{3}{2})^5=656100$ and
$6!\,5!\,(\frac{4}{2})^5=2764800$, 
and that we never store the leaves of the tree, which
implies that we could store at most around $500000$ positions, and we
store much less than that, as many generated positions are identical.
As $500000$ is almost $2^{19}$, $n=21$ is overkill anyway.

There is however a lesson to remember here: despite what many people say or write, the larger
is not always the better for hash tables. Sometimes, you first have to
keep the hash in the cache. 

\subsection{The breadth first algorithm}
Implementing the breadth first algorithm is not much more complicated
than implementing the depth first algorithm.
We first need to create a data structure that contains the information
needed to build the numbers generated by a subset of the initial
pool. For example, we need to know how to build the numbers generated
by the first, second and fourth number of pool. In order to do this
efficiently, we create an array of list where the $i$-th element
contains the list of pairs of sets to combine in order to build the numbers
generated by the subset represented by the binary
decomposition of $i$. This might sound complicated, but is easy to
understand with a few examples:
\begin{itemize}
\item For $i=16$ we have $i=16=10000_2$, so this element will
just point to the fifth element in the initial pool of numbers.
\item The element at $i=5=101_2$ points to the list of pairs of sets
  to combine. Here, we have to combine with the four operations the first
  element and the third element of the original pool, so there is
  only one pair   $(1,3)$. 
\item The element at $i=25=11001_2$ will contain the pairs $(1,24)$,
  $(8,17)$ and $(9,16)$ because to have all elements generated by the
  first, the fourth and the fifth element of the original pool we have
  to combine with the four operations (a) all elements generated by the
  fourth and the fifth with the first element, (b) all elements
  generated by the first and the fourth with the fifth element and (c)
  all elements generated by the first and the firth with the fourth
  element. 
\item The element at $i=57=111001_2$ will contain:
  \begin{itemize}
  \item the pairs $(1,56)$,
  $(8,49)$, $(16,41)$ and $(32,25)$ which are the numbers generated by
    the subsets generated by three elements to combine with the numbers
    generated by the subsets of one element 
  \item the pairs $(9,48)$, $(17,40)$, $(33,24)$ which are the
    numbers generated by the subsets generated by two elements
    combined with the other numbers generated by the subsets of the
    complementary two elements subsets.
    \end{itemize}
\end{itemize}
This array of list of pairs can be pre-computed and stored once and
for all. The size of the array is $2^n-1$ where $n=6$, so the array here
has 63 lists of pairs.

The rest of the algorithm is straightforward. Another array of the
same size is used, where the $i$-th element is an array
that will hold all numbers generated for the $i$ index. 

Let's see that on an example. If the initial pool of numbers is
$\{7,8,9,10,25,75\}$ we first copy 7 at position 1, 8 at position 2, 9
at position 4, 10 at position 8, 25 at position 16 and 75 at position
32. Then, all elements with an index having only 1 bit are
filled. Then we fill all elements having an index with 2 bits.
For example, element $3=11_2$ is $\{7+8,8-7,7\times8\}=\{15,1,56\}$,
element $5=101_2$
is $\{7+9,9-7,9\times7\}=\{16,2,63\}$, element $6=110_2$ is
$\{17,1,72\}$, and so on. When all elements with a 2-bits index are filled,
elements with a 3-bits index are filled. For example element $7=111_2$
is
$\{15+9,15-9,15\times9,1+9,9-1,56+9,56-9,56\times9\} \cup
\{16+8,16-8,16*8,16/8,2+8,8-2,8\times2,8/2,63+8,63-8,63\times8\} \cup
\{17+7,17-7,17\times7,1+7,7-1,72+7,72-7,72\times7\} =
\{24,6,135,10,8,65,47,504,24,8,128,2,10,6,\linebreak16,4,71,55,504,24,10,119,8,6,79,65,504\}$.

There remains a few implementation details to solve. Whether it is
better to use an array of arrays or an array of 
sets is unclear. Both structures have their advantages and their
disadvantages. An array has an access time which is constant,
while inserting a new number in a set of size $n$ takes $\log n$
operations when
using a binary balanced tree structure for the set. However, when using
sets, duplicates numbers are never kept and there are lot of
duplicates: even in the simple example above, there are already many
of them in the 3-bits 7th element. Another (minor) advantage of the sets is
that they use exactly the right number of elements while the size of
arrays has to be pre-computed at allocation time; however this minor
point may be circumvented in different ways: first we know a quite
good estimate of the size of each array, as the $N(p)$ numbers
computed in section~\ref{breadcomp} are an upper bound of the size of
a $p$-bits array. Moreover, it is possible to break the (large) arrays into a
list of smaller arrays which are allocated when needed.

Last, but not least, it is important to notice that while all numbers
have to be generated (of course), numbers generated by the full set of
the original pool (the array element with all bits set to 1)
do not have to be stored, as they will never be re-used. It is an
extremely important optimization of the code, as they are, and by far,
the largest set.

Experimental results with $n=6$ are the following : the breadth first algorithm
with an array-array structures solve the 13243 instances in 53s, and
in 89s with an array-set structures. The results of all the algorithms
are summarized in table~\ref{table1}.
\begin{table}[ht!]
\begin{center}
	\begin{tabular}{|l|r|r|}
	\hline
	Algorithm &Total Time&By instance\\
	\hline
	\hline
	Depth first	& 160&12.10E-3  \\
	Depth first / hash   	&26& 1.96E-3\\
	Depth first / hash-set   	&67& 5.05E-3 \\
	Breadth first / arrays   	&53& 4.00E-3 \\
	Breadth first / sets   	&89& 6.72E-3 \\
	\hline
	\end{tabular}
  \caption{\label{table1} Comparison of the algorithms for $n=6$ and 13243 instances}
\end{center}
\end{table}
The most efficient algorithm for $n=6$ is the depth first algorithm with
standard hash tables. The worst is the basic depth first
algorithm. Results are in accordance with the complexity analysis done
in section~\ref{seccomp}, as breadth first search is more efficient than depth
first search without hash tables. However, it would be extremely
interesting to see what happens with higher values of $n$.

The depth first algorithm with hash tables is extremely
efficient. There are other programs available on the net which claim
to solve also the complete set of instances such as \cite{Fouquet10},
but in 60 days (!).

\section{Scaling things up}
Since its beginning in 1972, the numbers round of the Countdown game
has never evolved, while its sister game, the letters round, has
seriously changed, going from 7 letters in 1972 to 10 letters
today. In 1972, computers were enable to solve the numbers round; nowadays,
it is solvable in less than a millisecond. So, as
in many games where computers have become much better than human
beings, the interest for the game has faded. Moreover, the game by
itself is not very difficult on the average for human
beings\footnote{{\em They should change
  the random number
  thingy so it doesn't come up with a really easy target
  number, meaning the contestants
  sit there like stiffs for nearly 30 seconds~}\cite{Virtue14}.}

There are thus two questions: is it
possible to modify the game in order to turn it into a difficult thing
for a computer, and is it possible to turn it into a game more
difficult for the players without modifying it too much?

There are two ways to change the difficulty of the game. The first one
is to choose the target number based on the values in the number set,
or even to choose only a tuple (numbers set,target value), such as the
number of operations for finding the target with the given numbers set
is high.

\begin{figure}[ht!]
  \begin{center}
    \includegraphics[width=7cm]{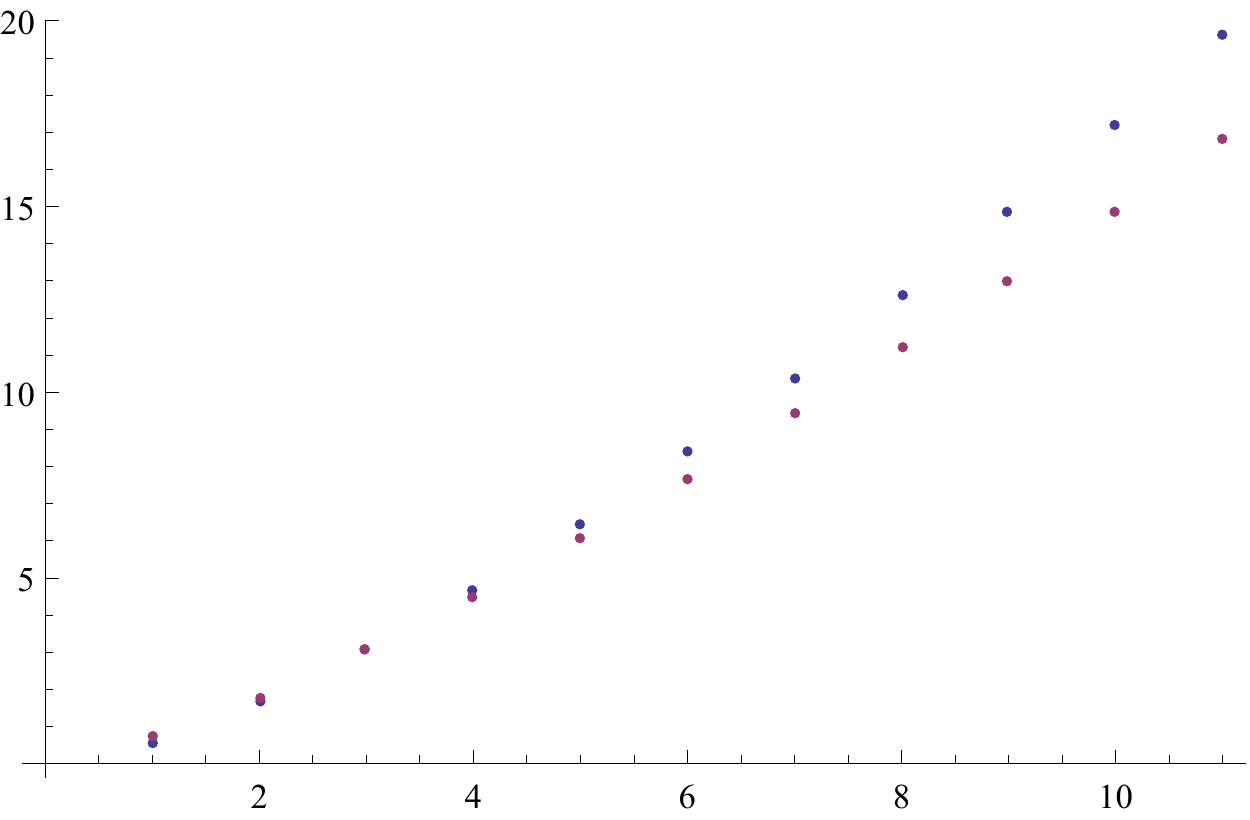}
    \caption{Complexity comparison. Blue: depth first. Red: breadth
      first\label{fig1}} 
  \end{center}
\end{figure}

The other idea comes from the complexity study which provides a hint: when the
size of the sets of available numbers increases, the game becomes
apparently extremely difficult. 
If we use the complexity formulas of sections \ref{secdepth} and
\ref{breadcomp}, we plot (figure~\ref{fig1}) in blue the
$\log_{10}$ of the number of operations required by the depth
first algorithm and in red the same quantity for the breadth first
algorithm. 
The breadth first algorithm quickly becomes {\bf much} more efficient 
than the depth first algorithm. However, its space complexity is
also increasing at almost the same rate as its time complexity, while
the space complexity of the depth first algorithm remains extremely
small. But these results do not take into account the hash table
effect for the depth first algorithm, or the set effect for the
breadth first algorithm, which are both going to become primary factors as
the number of duplicate positions and numbers will be much
more important as there will be much more ways to compute numbers
(especially small numbers) with a larger set of initial
numbers. The number of generated numbers is also going to increase: this
means that to have a depth first algorithm efficient, the size of
the hash tables has to be increased, which will take us out of the L2
and the L3 cache, and thus slow down significantly computations.

\begin{figure}[ht!]
  \begin{center}
    \includegraphics[width=7cm]{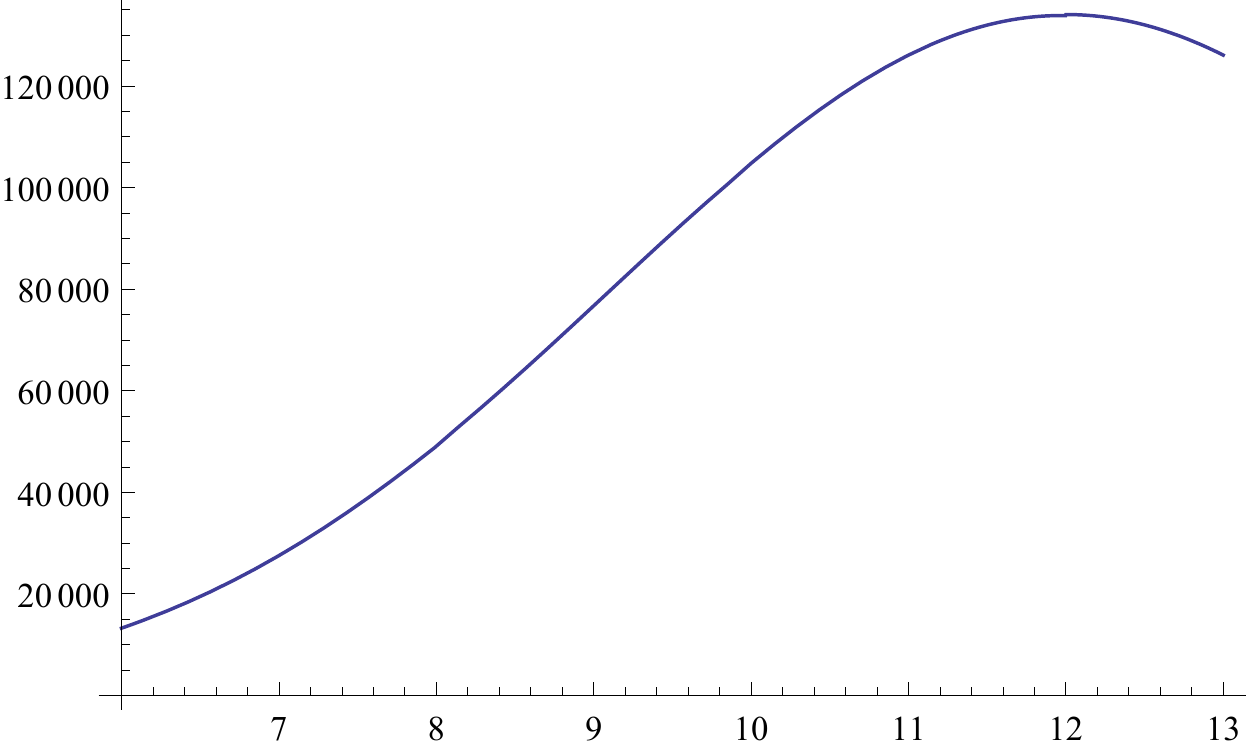}
    \caption{Number of instances ($n=6$)\label{figins}}
  \end{center}
\end{figure}

To compute the total number of different instances\footnote{The number
  of possible instances is not an indicator of the difficulty 
of the game, but we need these numbers in the next sections.}, we can extend
the formula in section~\ref{implem}:
\begin{eqnarray*}
&\binom{14}{n}&\mbox{with no pair}\\
+&\binom{10}{1}\times\binom{13}{n-2}&\mbox{with one pair}\\
+&\binom{10}{2}\times\binom{12}{n-4}&\mbox{with two pairs}\\
+&...&\\
+&\binom{10}{i}\times\binom{14-i}{n-2i}&\mbox{with $i$ pairs}\\
+&...&\\
+&\binom{10}{E(n/2)}\binom{14-E(n/2)}{n-2E(n/2)}&\mbox{with $E(n/2)$ pairs}\\
=&\sum_{i=0}^{E(n/2)} \binom{10}{i}\times\binom{14-i}{n-2i}&
\end{eqnarray*}
This formula is valid for $n\leq20$ and the number of instances is
$n(7)=27522$, $n(8)=49248$, $n(9)=76702$, and $n(10)=104753$. 
The results are summarized on figure~\ref{figins}.

\subsection{Solving for $n=6$}
\subsubsection{Standard game}
For $n=6$ we have 13243 possible sets. In the standard numbers round of
the countdown game,
we search for numbers in the range 101--999, so there are
$899\times13243=11905457$ possible problems. In table~\ref{tabledists} we
have the distance to the closest numbers: 10858746 games are solvable
(91.2\%), 743896 problems (6.25\%) have a solution at a distance of 1
(the nearest number).
\begin{table}[ht!]
\begin{center}
	\begin{tabular}{|l|r|r|r|}
	\hline
	distance&solved&\%solved&cumulative\\
	\hline
	\hline
        0&10858746&91.21\%&91.21\%\\
        1&743896&6.25\%&97.46\%\\
        2&100517&0.84\%&98.30\%\\
        3&36186&0.30\%&98.60\%\\
        4&19387&0.16\%&98.76\%\\
	\hline
	\end{tabular}
  \caption{\label{tabledists}Distance to the solution}
\end{center}
\end{table}

1226 instances out of 13243 (9.2\%) solve all 
target numbers in the range 101-999. One instance ($\{1,1,2,2,3,3\}$) solves
none. 
\begin{figure}[ht!]
  \begin{center}
    \includegraphics[width=7.5cm]{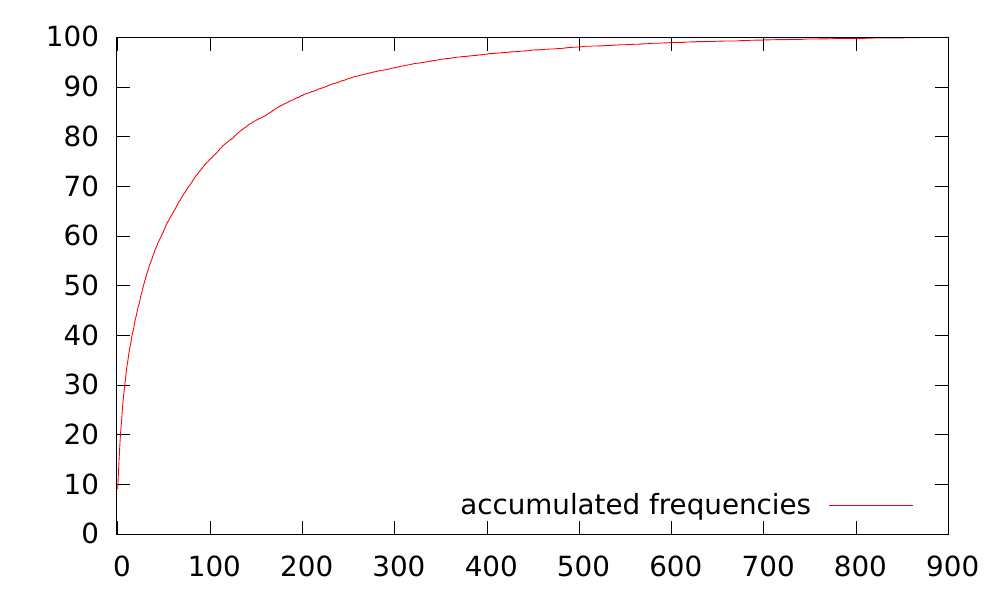}
    \caption{x-axis: number of numbers not found, y-axis:percentage of
      instances\label{figtirages6}}
  \end{center}
\end{figure}
On figure~\ref{figtirages6}, we see that 9998 (75.5\%) of the possible
instances solve all possible games with less than 100 numbers missing
in the range 101--999. 

The easiest numbers to find are 102, 104 and 108 which are found by
13240 instances (99.98\%). The most difficult number to find is 947,
which is only found by 9017 instances (68\%).
\begin{figure}[ht!]
  \begin{center}
    \includegraphics[width=7.5cm]{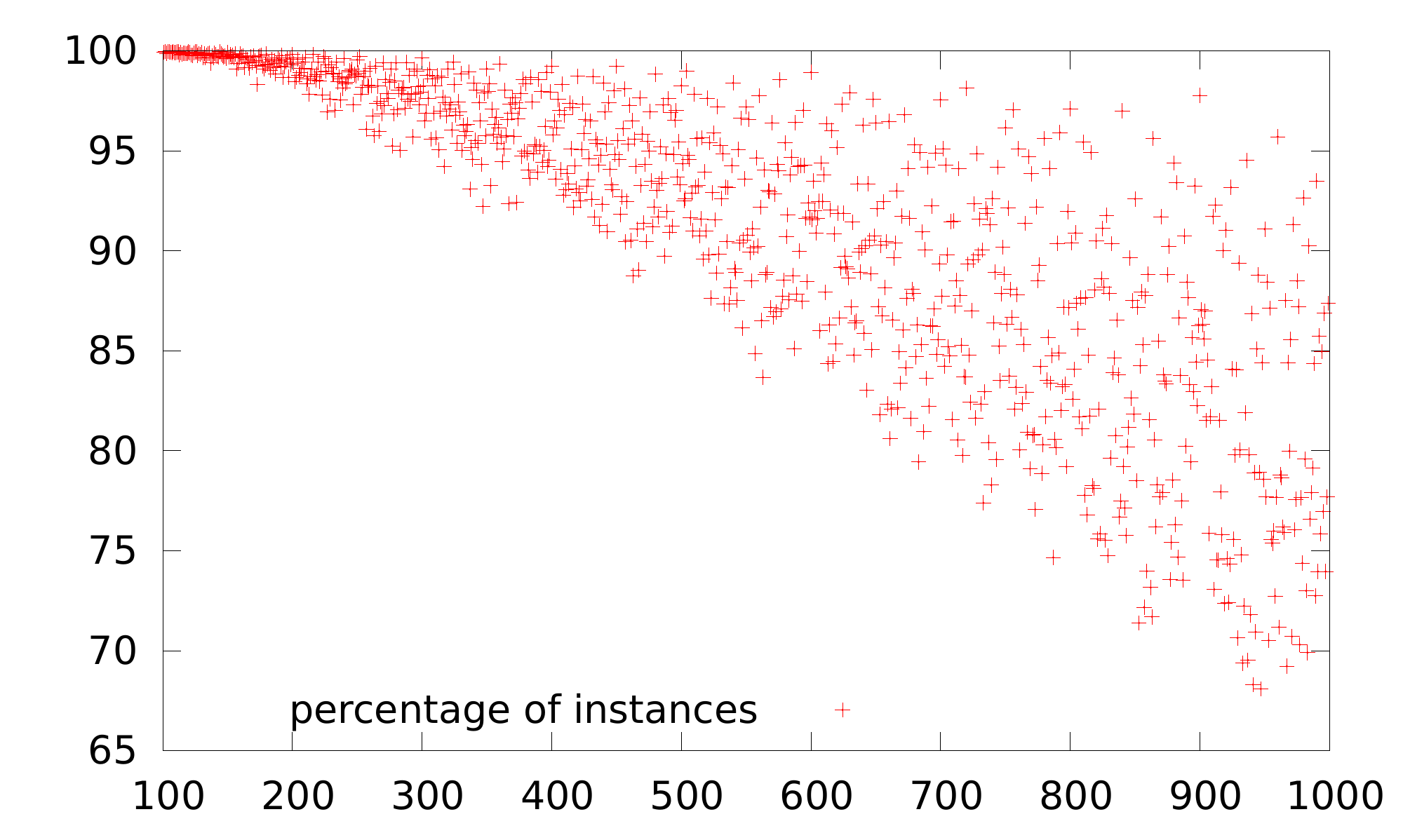}
    \caption{x-axis: number to find, y-axis:percentage of
      instances finding this number\label{fignumbers6}}
  \end{center}
\end{figure}
On figure~\ref{fignumbers6}, we see that, as we might have expected,
the easiest numbers to find are the lowest, and the most difficult are
the highest. Numbers below 300 are all found by 95\% of the possible
instances. 

Another interesting statistic for the player of the British version of
the game is how the distribution of large (25, 50, 75, 100) and small
(1 to 10) numbers influence the number of solutions available. Each large
number is present in
$\binom{13}{5}+\binom{10}{1}\binom{12}{3}+\binom{10}{2}\binom{11}{1}=3982$
instances (30\%) of the 13243 instances, while 
each small number appear in
$\binom{13}{5}+\binom{13}{4}+(\binom{10}{1}-1)\binom{12}{3}+\binom{9}{1}\binom{12}{2}
+(\binom{10}{2}-\binom{9}{1})\binom{11}{1}+\binom{9}{2}=5008$ instances
(38\%). 

\begin{figure}[ht!]
  \begin{center}
    \includegraphics[width=7.5cm]{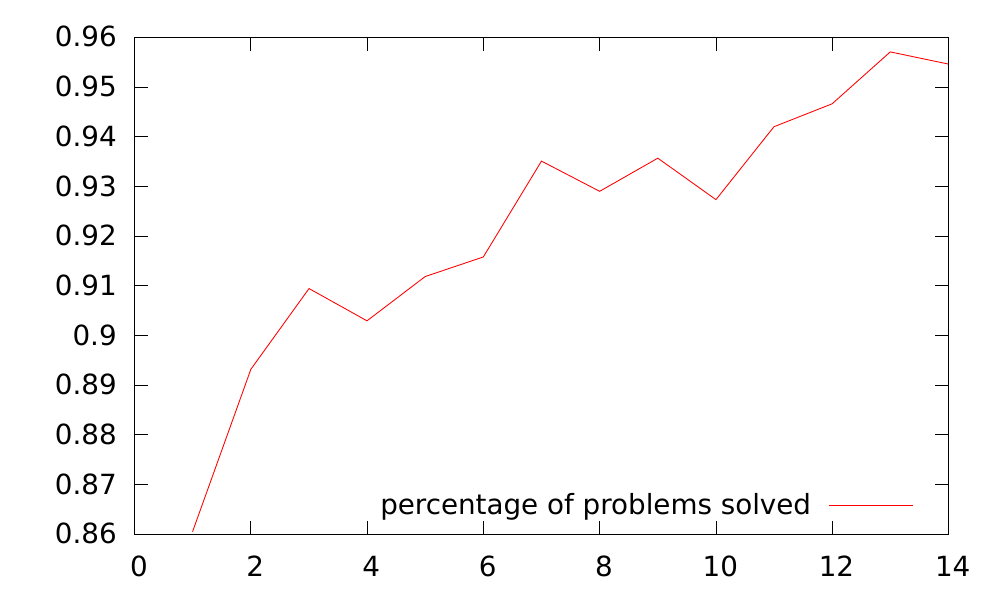}
    \caption{y-axis:percentage of
      problems solved when number $x$ is in the set
      ($11\leftrightarrow25$, $12\leftrightarrow50$, $13\leftrightarrow75$,
      $14\leftrightarrow100$)
      \label{figsolved6}}
  \end{center}
\end{figure}

On figure~\ref{figsolved6}, we see the percentage of problems solved
when a number $x$ is in the original set. 
The worst number is 1 (only 86\% problems are solved when 1 is in the set) and
the best is 75 (almost 96\% are solved when 75 is in the set). But the
differences are not that important between large and small numbers:
for $x=9$, 93.5\% are solved, not that far from the 94.2\% for $x=25$.

In table~\ref{tablesolved}, we see how the number of large numbers in
the set influences the resolution. For example there are
$\binom{10}{6}+ \binom{10}{1}\binom{9}{4}+
\binom{10}{2}\binom{8}{2}+ \binom{10}{3}=2850$ instances with no large
numbers, and thus $2850*899=2562150$ problems and 1963762 of
these problems can be solved (77\% success rate).
\begin{table}[ht!]
\begin{center}
	\begin{tabular}{|l|r|r|r|}
	\hline
	nb large &problems&solved&\%solved\\
	\hline
	\hline
        0&2562150&1963726&77\%\\
        1&5221392&4966076&95\%\\
        2&3317310&3192103&96\%\\
        3&755160&693131&92\%\\
        4&49445&43710&88\%\\
\hline
        &11905457&10858746&91\%\\
	\hline
	\end{tabular}
  \caption{\label{tablesolved} Percentage of instances solved as a
    function of the number of large numbers in the set}
\end{center}
\end{table}
The influence of large numbers is much more visible here. Instances
with 1--3 large numbers have a success rate of 92--96\%, and even
with the four large numbers (25,50,75,100) the success rate is higher
that with none of 
them. However, the importance of large numbers must not be
overestimated, as \cite{Crosswords} does. The 4-tuple (25,50,75,100)
has a success rate of 88\%,
much less than (5,7,9,100) which has a success rate of 99.86\% and
contains only one large number (the worst 4-tuple is (1,1,2,2) with a
success rate of 37\%).

There is another site (\cite{Kitsune12}) in french which
advertises the kitsune program and gives some stats. However, it takes
a few hours to compute them, while this program takes only a few seconds.
So, for the fans of statistics and results, here are some other ``funny''
facts: 
\begin{itemize}
\item The best 3-tuple is (7,9,100) with a success rate of 99\%, the
  worst is (1,1,2) (58\%). The
  best pair is (7,100) (97.7\%), the worst (1,1) (73\%). The worst
  5-tuple is
  (1,1,2,2,3) (14\%). There are 7
  5-tuples which have a 
  success rate of 100\% (any number can be added to any of these
  5-tuples, and the resulting set will solve the 
  899 problems): (4,6,7,9,100), (2,5,8,9,100), (2,5,6,9,100), 
  (5,6,7,9,100), (4,7,9,10,100), 
  (2,7,9,10,100), (2,4,7,9,100). However an additional number {\bf is}
  needed.  
\item No five numbers set can solve
  by itself all the problems. $\{4,6,7,9,100\}$ and
  $\{2,5,8,9,100\}$ solves 753 out of 899, $\{2,5,6,9,100\}$ solves
  751. $\{2,3,8,9,100\}$ is the next best with 748 solved but it
  doesn't appear in the list of the best 5-tuples.
\item The success rate drops quickly with the size of the
  set. With four numbers sets, the best we can get is $\{2,5,8,100\}$
  which solves only 159 problems.
\item There is no instance with four large numbers which solves all
  problems
\item There are five instances with only small numbers which solve all
  problems, the ones with the least sum (41) being
  $\{2,5,7,8,9,10\}$ and $\{3,4,7,8,9,10\}$
\item There is no instance solving all problems with all numbers less
  or equal to 9.
\item The instance with the largest weight (244) solving all problems is 
$\{2,8,9,50,75,100\}$ It is also the only instance containing 50, 75 and
  100 that solves all problems.
\item There is only one instance solving all problems with all numbers
  greater or equal than 8: $\{8,9,9,10,25,75\}$
\item the problem which requires the largest intermediate result is 
$\{3,3,25,50,75,100\}$ and 996, with $50+3=53, \, 53\times25=1325,\,
  1325+3=1328,\, 1328\times75=99600,\, 99600/100=996$ Thus
  programs using only short unsigned integers (up to 65535) could not
  solve all problems.
\end{itemize}

\subsubsection{Selecting problems}
As all instances have been solved, we have a complete database;
for a given number set and a given target number we know if it can be
solved and how many operations are necessary to solve it, or
how close is the nearest findable number
when it can't be solved. With this database, it is extremely
easy to select only interesting problems. There can be many different
selection criteria: solvable problems requiring more than 4 (or 
5...)
operations, or unsolvable problems with the nearest number at a
minimal given distance, or unsolvable problems with the nearest
number requiring more than 4 operations, etc\ldots
This would turn the number round in something worth
watching again.
\label{selection}

\subsubsection{Using a larger set to pick numbers}
Another way to make the game harder would be to use all available
numbers between 1 and 100 when picking the set.
Building the full database is much more computing intensive. In the
standard game we have 13243 sets, when picking $k$ numbers between 1
and $n$ (including repetitions) we have
$\binom{n+k-1}{k}=\binom{100+6-1}{6}=1609344100\simeq1.6\,10^9$
possible sets, and $1446800345900\simeq1.4\,10^{12}$ problems. 
Building the database took 12 hours on the cluster described in
section~\ref{implem}.

Table~\ref{tabledists2} gives the distance to the solution, as
table~\ref{tabledists} does for the standard game.
Percentages are similar to the standard problem.
\begin{table}[ht!]
\begin{center}
	\begin{tabular}{|l|r|r|r|}
	\hline
	dist&solved&\%solved&cumulative\\
	\hline
	\hline
        0&1329106855477&91.86\%&91.86\%\\
        1&105091143229&7.26\%&99.12\%\\
        2&8508187551&0.59\%&99.71\%\\
        3&2112923902&0.14\%&99.85\%\\
        4&808768195&0.06\%&99.91\%\\
	\hline
	\end{tabular}
  \caption{\label{tabledists2}Distance to the solution}
\end{center}
\end{table}

\begin{figure}[ht!]
  \begin{center}
    \includegraphics[width=7cm]{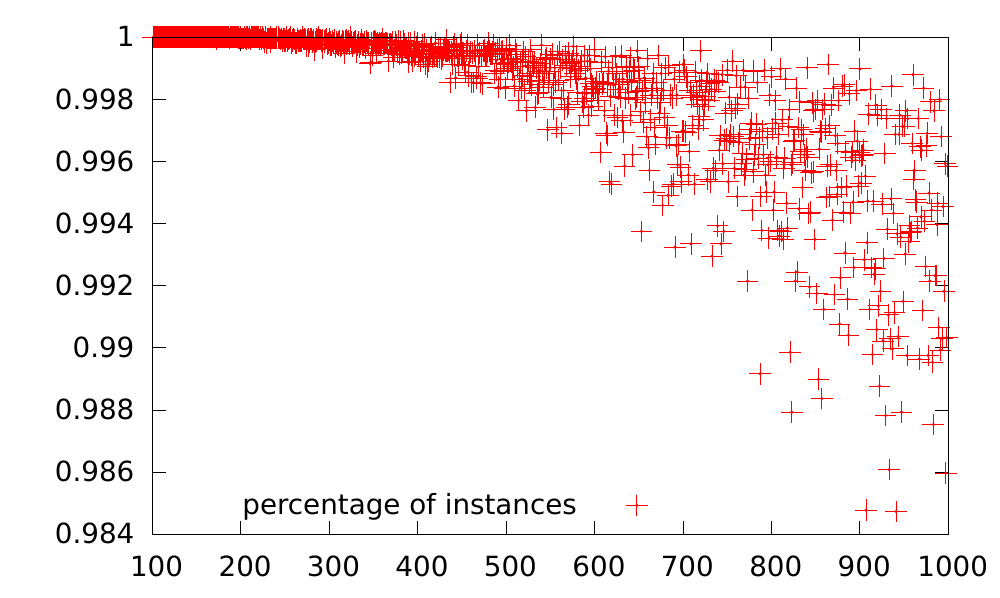}
    \caption{x-axis: number to find, y-axis:percentage of
      instances finding this number($n=6$, extended set)\label{fignumbers6b}}
  \end{center}
\end{figure}

In figure~\ref{fignumbers6b}, we have the same results as in
figure~\ref{fignumbers6}. Percentages are higher which means that on the
average, the problem is easier to solve with numbers picked randomly
between 1 and 100.

There are 73096123 (4.5\%) sets that solve all problems. This is less
in percentage ($1226/13243\simeq9.2\%$) than for the standard game,
but there are 60000 times more sets if we consider the raw numbers. 
So we can select some sets with 
specified characteristic that would make them difficult
for human beings, while maintaining the diversity of the problem.
There are for example 52253 sets that solve all
problems while being composed only by prime numbers, 48004 by primes $\geq 2$,
22136 by primes $\geq 3$, 8912 by primes $\geq 5$, 4060 by primes $\geq
7$, 1526 by primes $\geq 11$, 500 by primes $\geq 13$, 132 by primes
$\geq 17$, and 4 by primes $\geq 23$. As incredible as it might look,
the set $\{23,29,31,37,43,61\}$ solves the 899 problems.

Another
criteria could be to select sets where all numbers are greater than a
given one; there are for example 20602 sets with all numbers $>25$
that solve the 899 problems. The set $\{35,37,38,43,45,59\}$ is one of
them\ldots This method can be combined with the one described in
section~\ref{selection}, by
choosing only target numbers that require a minimum number of
operations. Here again, the possibilities are endless, and it would
turn the numbers game into something really difficult while always
using 6 numbers.

\subsection{Solving for $n=7$}
Using equations \ref{dmax}, \ref{dmin}, \ref{bmax} and \ref{bmin} we
find that the maximal and minimal number of operations for the depth first
algorithm are 
$d_{max}(7)=232 243 200$ and $d_{min}(7)= 41 334 300$. For the breadth first
algorithm, we have 
$b_{max}(7)= 49 951 531$ and
$b_{min}(7)=  9 379 195$. We thus have
$\frac{d_{max}(7)}{d_{max}(6)}=84$,
$\frac{d_{min}(7)}{d_{min}(6)}=63$,
$\frac{b_{max}(7)}{b_{max}(6)}=43$,
$\frac{b_{min}(7)}{b_{min}(6)}=32$.

In table~\ref{table7} we have the results of the experimentation with
the five algorithms with $n=7$. 
\begin{table}[ht!]
\begin{center}
	\begin{tabular}{|l|r|r|}
	\hline
	Algorithm &Total Time&By instance\\
	\hline
	\hline
	Depth first	& 740&740E-3  \\
	Depth first / hash   	&36& 36E-3\\
	Depth first / hash-set 	&114& 114E-3 \\
	Breadth first / arrays 	&109& 109E-3 \\
	Breadth first / sets   	&131& 131E-3 \\
	\hline
	\end{tabular}
  \caption{\label{table7} Comparison of the algorithms for $n=7$ and
    1000 instances}
\end{center}
\end{table}

We see that with the depth first algorithm, the time for solving
instances with 7 numbers is 62 ($740/12$) times larger than with
$n=6$. This is completely compatible with the minimal complexity of
this algorithm, which predicts a ratio of 63.

With the breadth first algorithm, the time for solving
instances with 7 numbers is 28 ($109/4$) times larger than with
$n=6$. This is slightly less than what was expected (a ratio of 32)
but remains in line with what was expected.

With the depth first algorithm with hash table, the ratio is only
18. Hash tables are getting more and more efficient, as small numbers
are generated more often. An analysis of the optimal size of the hash
table shows that the best size is around $2^{19}$ instead of $2^{15}$
for 6 numbers:
more space is needed to hold more numbers, even if data can not remain
inside the L2 cache.

\begin{figure}[ht!]
  \begin{center}
    \includegraphics[width=7.5cm]{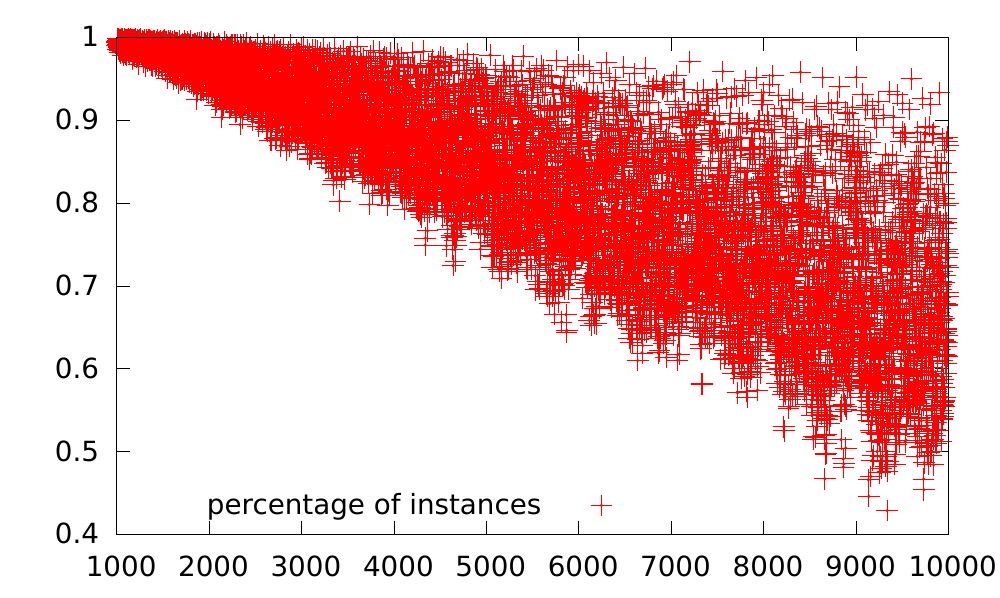}
    \caption{x-axis: number to find, y-axis:percentage of
      instances finding this number for $n=7$\label{fignumbers7b}}
  \end{center}
\end{figure}

Regarding the resolution of problems we see on
figure~\ref{fignumbers7b} how numbers are found. 
With an extra number
in the set, the success rate becomes extremely high. All numbers are
found by at least 98.5\% of the instances: the problem has become too
easy. 

\begin{figure}[ht!]
  \begin{center}
    \includegraphics[width=7.5cm]{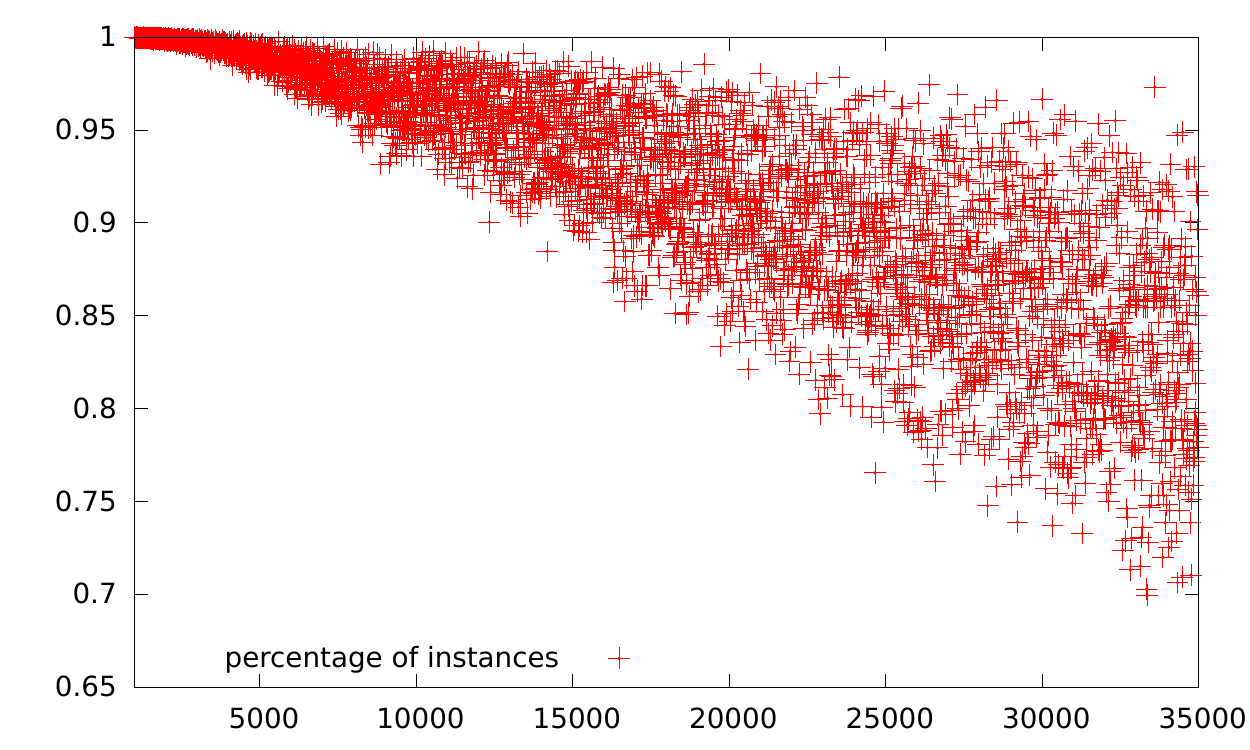}
    \caption{x-axis: number to find, y-axis:percentage of
      instances finding this number for $n=7$\label{fignumbers7}}
  \end{center}
\end{figure}

If we try to find numbers in the range 1000--10000 instead of
100--1000, we see on figure~\ref{fignumbers7} that the problem is now
too difficult.
The right solution is to look for numbers in the range 1000--6000 (the
most difficult number to find is then 5867, with  65\% instances
finding it) . The
success rate is now almost the same as what it was with 6 numbers in
the range 100--1000, but with a resolution time which is 20 times
higher. However, the solution of a problem is found by the best
algorithm in 36 milliseconds, which is still much too fast to put the
machine in the same league as a human being\ldots

\subsection{Solving for $n=8$ }
We have here
$b_{min}(8)=   363 099 899$, 
$d_{min}(8)= 3 472 081 200$ and
$\frac{d_{min}(8)}{d_{min}(7)}=84$ and 
$\frac{b_{min}(8)}{b_{min}(7)}=39$.

In table~\ref{table8} we have the results of the experimentation with
the five algorithms with $n=8$. 
\begin{table}[ht!]
\begin{center}
	\begin{tabular}{|l|r|r|}
	\hline
	Algorithm &Total Time&By instance\\
	\hline
	\hline
	Depth first	& 610&61  \\
	Depth first / hash   	&12&1.2 \\
	Depth first / hash-set 	&37&3.7  \\
	Breadth first / arrays 	&44& 4.4 \\
	Breadth first / sets   	&41&4.1  \\
	\hline
	\end{tabular}
  \caption{\label{table8} Comparison of the algorithms for $n=8$ and
    10 instances}
\end{center}
\end{table}

We see that with the depth first algorithm, the time for solving
instances with 8 numbers is 83 ($61/0.740$) times larger than with
$n=7$. This is completely compatible with the minimal complexity of
this algorithm, which predicts a ratio of 84.

With the breadth first algorithm, the time for solving
instances with 8 numbers is 40 ($4.4/0.109$) times larger than with
$n=7$. This is exactly what was expected (a ratio of 39).
We notice that the breadth-first-set algorithm is becoming faster than the
breadth-first-array algorithm. With 8 numbers, we are generating more
and more small duplicate numbers, and thus the time lost with the
$\log n$ access for sets is now compensated by the time gained with
the elimination of all these duplicates numbers. 
With the depth first algorithm with hash table, the ratio is 
33. The memory needed to run the breadth-first-array algorithm is
1.5Gb. The breadth-first-set algorithm still has small memory
requirements. For
the depth-first with hash, the optimal value of the size of the hash
table is around $2^{23}$ elements.

\begin{figure}[ht!]
  \begin{center}
    \includegraphics[width=7.5cm]{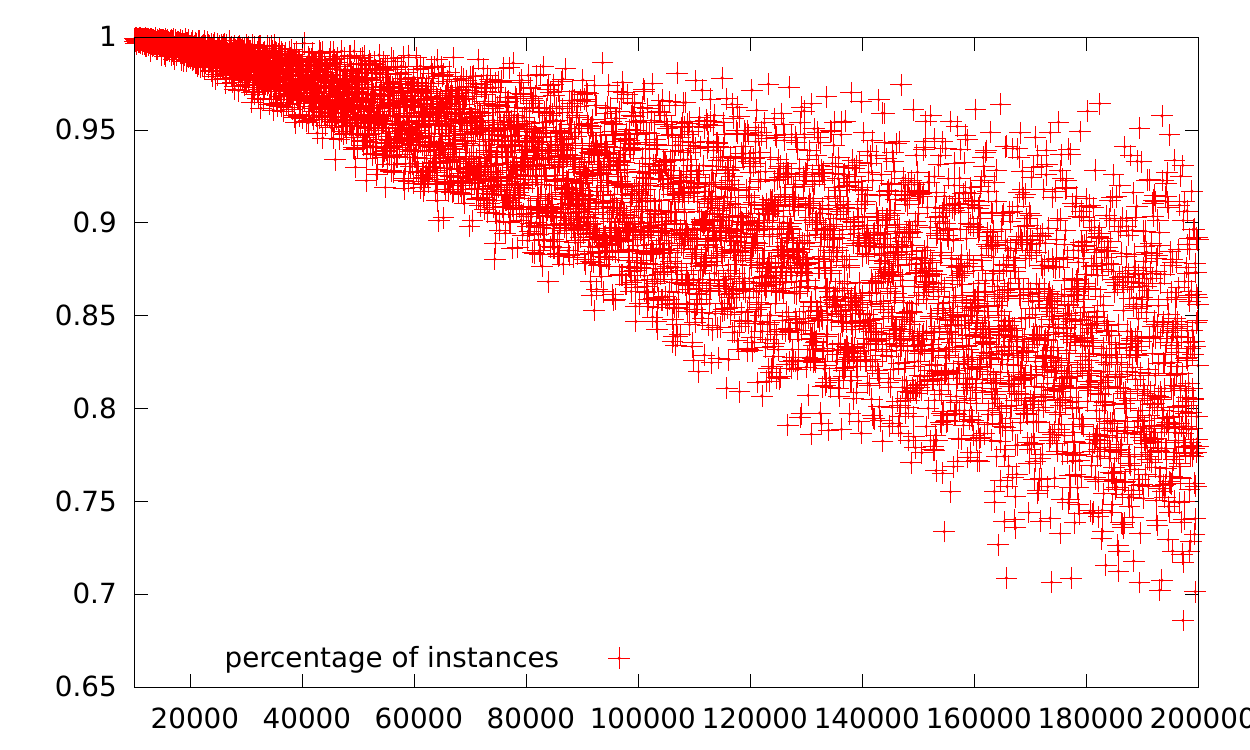}
    \caption{x-axis: number to find, y-axis:percentage of
      instances finding this number for $n=8$ (sampled every 9
      points)\label{fignumbers8}}
  \end{center}
\end{figure}

The results are presented in figure~\ref{fignumbers8}. Computation
took a few hours.
There again, with an additional number, the problem becomes too easy
to solve in the previous range (1000--10000). The correct range 
must be extended up to 35000 as we have then roughly the same mean success
rate as with the 
standard game (the most difficult number to find is 34763 with a
success rate of 66\%). However, if
the depth first program is now unable to 
compute the solution in less than 30s (English game) or 45s (french
game), the depth-first with hash still finds a solution in 1.2s on the
average.

\subsection{Solving for $n=9$}
We have
$\frac{d_{min}(9)}{d_{min}(8)}=108$ and 
$\frac{b_{min}(9)}{b_{min}(8)}=48$.
Thus the standard depth first algorithm should take more
than 6000s to solve a single instance and the breadth first
algorithm with arrays should need around 40Gb of memory, that the
computer used for these tests don't have.

In table~\ref{table9} we have the results of the experimentation with
three algorithms with $n=9$. The depth first algorithm wasn't, as
expected, able to 
solve even a single instance in less than 1 hour. The breadth first
algorithm with arrays generated an ``Out of memory'' error.
\begin{table}[ht!]
\begin{center}
	\begin{tabular}{|l|r|r|}
	\hline
	Algorithm &Total Time&By instance\\
	\hline
	\hline
	Depth first	&- & - \\
	Depth first / hash   	&147&14.7 \\
	Depth first / hash-set 	&543&54.3  \\
	Breadth first / arrays 	&-& - \\
	Breadth first / sets   	&467&46.7  \\
	\hline
	\end{tabular}
  \caption{\label{table9} Comparison of the algorithms for $n=9$ and
    10 instances}
\end{center}
\end{table}

\begin{figure}[ht!]
  \begin{center}
    \includegraphics[width=7.5cm]{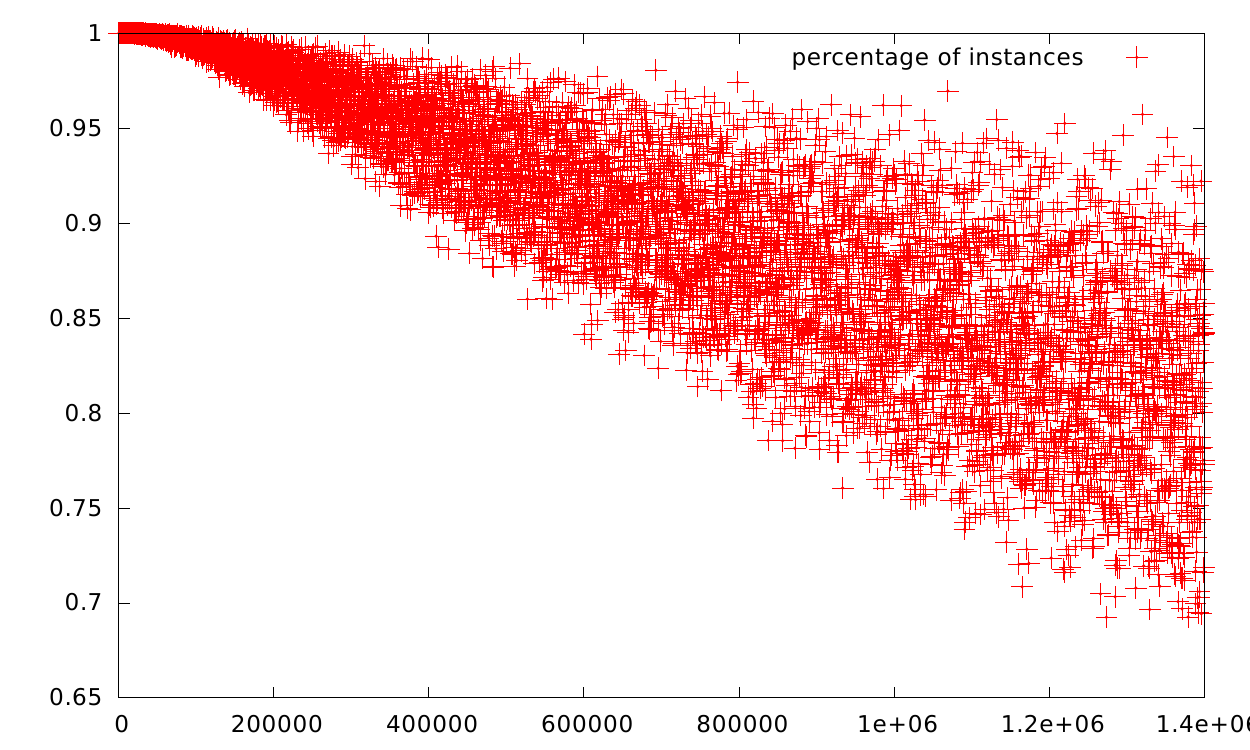}
    \caption{x-axis: number to find, y-axis:percentage of
      instances finding this number for $n=9$ (sampled every 37
      points)\label{fignumbers9}}
  \end{center}
\end{figure}

The results are presented in
figure~\ref{fignumbers9}. 
We have to extend the range up to around 200000 (the most difficult
number to find is 190667 with a success rate of 66\%).
Computing complete results took 3 days.

\subsection{Solving for $n=10$}
For $n=10$ we are at last entering uncharted territory. The average
time to solve one instance of the problem seems to be around 1 to 3
minutes, so it seems impossible to use an exhaustive algorithm. We are
at last back in the heuristics land.

\begin{figure}[ht!]
  \begin{center}
    \includegraphics[width=7.5cm]{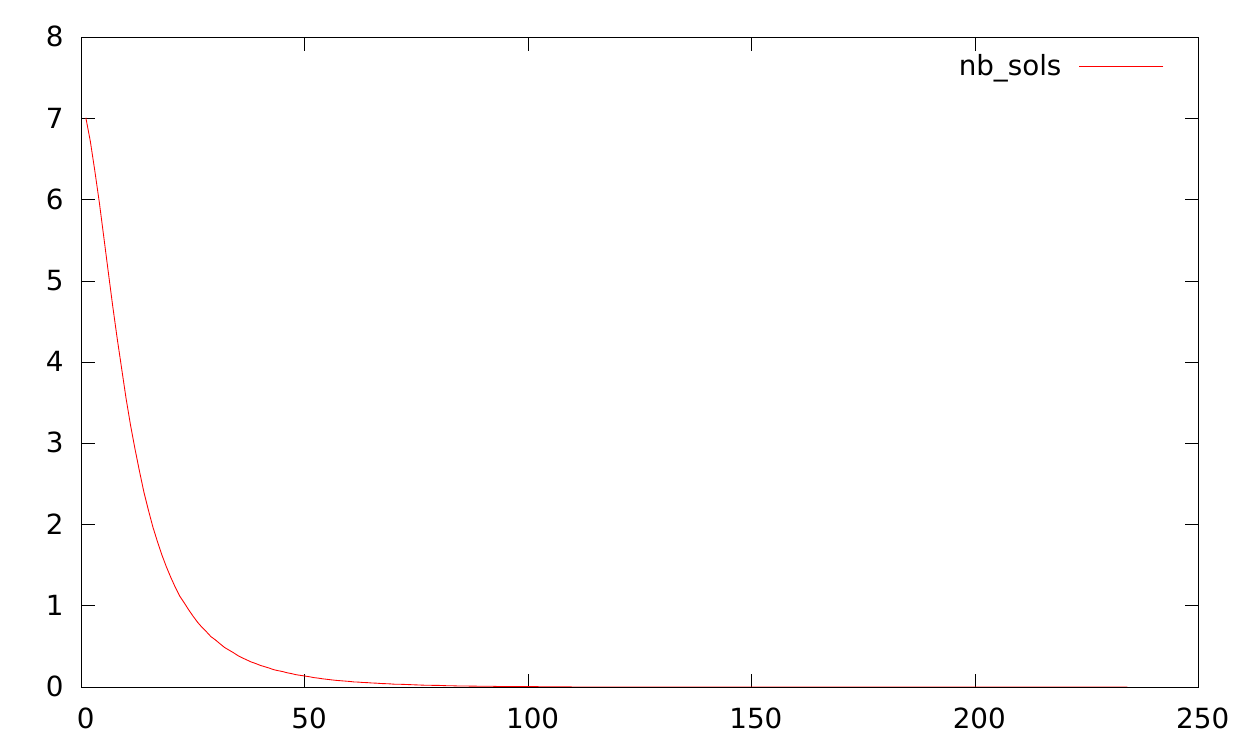}
    \caption{x-axis: number to find, y-axis:percentage of
      instances finding this number for $n=10$ (sampled every 509
      points)\label{fignumbers10}}
  \end{center}
\end{figure}

The results are presented in
figure~\ref{fignumbers10}. 

Complete results were computed in 20 hours on the cluster described in
section~\ref{implem}. Some pools such as
$\{5,6,7,8,9,10,25,50,75,100\}$ took more than one hour to complete. We
had to extend the range over 1000000 
to have similar results regarding success rate (up to 1000000 the most
difficult number to find is 986189 with a 67\% success rate). 

\section{A slightly modified problem}
The problem is easy to solve because it is a finite one: at each
step, the set of available numbers is reduced by one unit, and thus
any computer program can solve it even with a very large set of
numbers. An other solution to turn the game into a more interesting
one would be
to add a simple operation: the possibility to replace any available
number by its square.

Let's see this on an example: how to find 999 using
\{1,2,3,4,5,6\}. This is an unsolvable problem without the square
operation, but it is now not the case anymore:
{\small
\begin{verbatim}
   Operations     Remaining
  3 x  6 =  18  {1,2,4,5,18}
 18 x 18 = 324  {1,2,4,5,324}
  4 +  5 =   9  {1,2,9,324}
324 +  9 = 333  {1,2,333}
  1 +  2 =   3  {3,333}
333 x  3 = 999  {999}
\end{verbatim}
}

This modification changes the nature of the game, because it is not
any more a ``finite'' one, at least in theory. Thus, we can
have long and complex computations to find results.
Let's see it on an example: how to find 862 using the
\{1,10,10,25,75,100\} set. The shortest computation requires 14 steps
(while in the standard game we can never have more than 5 steps) and
uses very large numbers:
{\small
\begin{verbatim}
{1,10,10,25,75,100}
10 - 1 = 9         
{9,10,25,75,100}
100 x 100 = 10000
{9,10,25,75,10000}
9 x 9 = 81
{81,10,25,75,10000}
10 x 10 = 100
{81,100,25,75,10000}
100 x 100 = 10000
{81,10000,25,75,10000}
10000 + 10000 = 20000
{81,20000,25,75}
75 x 75 = 5625
{81,20000,25,5625}
5625 x 5625 = 31640625
{81,20000,25,31640625}
20000 x 20000 = 400000000
{81,400000000,25,31640625}
400000000 - 31640625 = 368359375
{81,368359375,25}
25 x 25 = 625
{81,368359375,625}
625 x 625 = 390625
{81,368359375,390625}
368359375 / 390625 = 943
{81,943}
943 - 81 = 862
\end{verbatim}
}

The program has to be slightly modified to include the possibility to
raise a number to its square at any time, and it must also be limited: we
have to set an upper bound $A$ above which we do not square numbers
anymore. Without this bound, the algorithm might not stop. Moreover,
because of implementation issues, the maximal value of $A$ that can be
tested with 64 bits arithmetic is 45000.

The possibility of squaring numbers seriously increases the complexity
of the program. As we are only interested in finding whether a given
set is able to solve all numbers in the range 101--999, we stop as
soon as all these numbers have been found and do not keep on searching
for the shortest solution available. With this optimization, and by
using all the other optimizations presented above, computation time is
not really an issue, at least for values of $A$ up to 50000.

We see in table~\ref{tableA} results for different values of $A$.
\begin{table}[ht!]
\begin{center}
	\begin{tabular}{|r|r|r|l|}
	\hline
	A &Sets&Instances&\% unsolved\\
	\hline
	\hline
	1       & 12017 &  1046711 & 8.79\% \\
	2	& 10757 &   758822 & 6.37\% \\
	3	&  9059 &   503409 & 4.22\% \\
	4	&  6275 &   196070 & 1.65\% \\
	5	&  5004 &   128631 & 1.08\% \\
	6	&  3507 &    74137 & 0.622\% \\
	7	&  2478 &    48932 & 0.411\% \\
	8	&  1637 &    29165 & 0.245\% \\
	9	&   926 &    13889 & 0.117\% \\
	10	&   593 &     7231 & 0.0607\% \\
	20	&   294 &     2706 & 0.0227\% \\
	30	&    99 &      443 & 0.00372\% \\
	40	&    55 &      311 & 0.00261\% \\
	50	&    41 &      225 & 0.00189\% \\
	60	&    40 &      221 & 0.00185\% \\
	70	&    35 &      206 & 0.00173\% \\
	80	&    31 &      166 & 0.00139\% \\
	90	&    28 &      130 & 0.00109\% \\
	100   	&    20 &       77 & 0.000647\% \\
	200   	&    17 &       62 & 0.000520\% \\
	300   	&    16 &       55 & 0.000461\% \\
	400   	&    16 &       54 & 0.000454\% \\
	2000   	&    16 &       54 & 0.000454\% \\
	3000   	&    15 &       53 & 0.000445\% \\
	4000    &    14 &       52 & 0.000437\% \\
	10000   &    14 &       51 & 0.000428\% \\
	20000   &    13 &       49 & 0.000412\% \\
	45000   &    13 &       49 & 0.000412\% \\
	\hline
	\end{tabular}
  \caption{\label{tableA} Number of sets having at least one instance
    not solved and unsolved instances as a function of
    $A$}
\end{center}
\end{table}

For $A=1$ the results are the
results of the standard algorithm, because squaring 1 gives 1: 
$1046711$ instances (of 11905457) are not solve, and there is at least
one number not found for $12017$ sets of numbers. The number of
unsolved instances reduces quickly in the beginning of the curves, but
then slows down. 

\begin{figure}[ht!]
  \begin{center}
    \includegraphics[width=7.5cm]{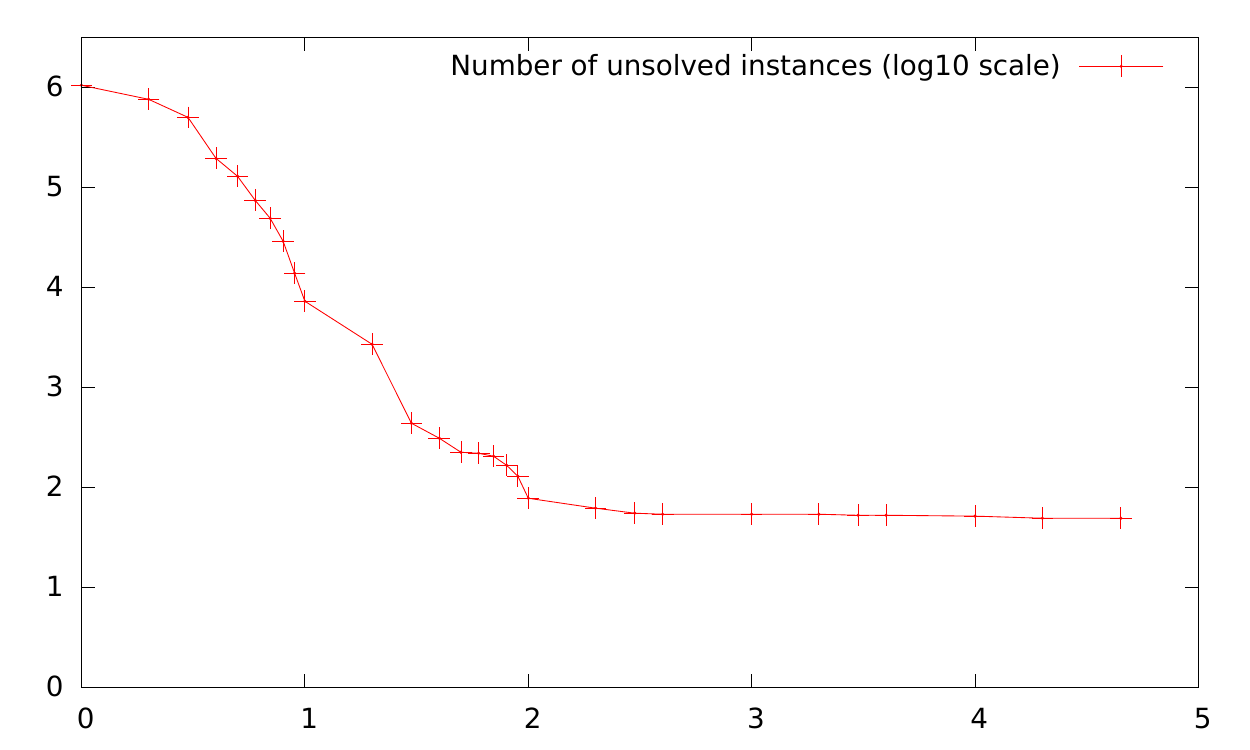}
    \caption{x-axis: $A$ ($\log_{10}$ scale), y-axis: number of unsolved
      instances ($\log_{10}$ scale)\label{fig13}}
  \end{center}
\end{figure}

The results are presented graphically in
figure~\ref{fig13}. 
The 49 instances not solved (with $A=45000$) are the following ones:
{\small
\begin{verbatim}
1 1 10 10 25 100: 858 
1 1 10 10 25  75: 863 
1 1 10 10 50 100: 433 453 547 683 773 
                  853 
1 1 10 10 50  75: 793 853 978 
1 1 10 10 75 100: 433 453 457 478 547 
                  618 653 682 708 718 
                  778 793 822 853 892 
                  907 958 978 
1 1 10 25 75 100: 853 863 
1 1 10 50 75 100: 793 813 853 978 
1 1  5  5 25 100: 813 953 
1 1  7  7 50 100: 830 
1 1  8  8  9   9: 662 
1 1  9 10 10 100: 478 573 587 598 
1 1  9  9 10 100: 867 
1 9  9 10 10 100: 867 947 957 958 967 
\end{verbatim}
}

If searching for results
in the range 1001--9999 instead of 101-999, the percentage of solvable
problems remains extremely high (at least 99.9705\%: only 35200
instances out of
119173757 seem to be unsolvable). 
However problems
are usually much more difficult for a human being, as they require
using much larger numbers. Another interesting proposal to revive
the current countdown game would be to keep on using 6 numbers drawn
in the same pool, but to search
now for numbers in the range 1001-9999 and to allow using the square operation.

From a theoretical point of view, 
the main question is: {\em are there some instances
that can never be solved whatever the value of $A$? }

This question is a complex one and requires further research: on the
one hand, we can hope that by searching with large enough values of
$A$ we would solve all instances of the problem. However, if our
search is not successful, it is pretty much unclear how to demonstrate
that a given instance has no solution. {\em This could indeed be an
  example of a simple undecidable problem.}

\section{A subsidiary problem: reducing the number of solutions}
When looking for a specific number there are often many different
ways to find the result. However, most of these solutions are
``identical'' from a human point of view. There are unfortunately no
clear boundary between ``identical'' and ``different'' solutions.
There are some elementary rules that can be used to reduce the number
of solutions, but it is highly improbable that the filtered solutions
would satisfy all fans of the game.

 We use
postfix notation ($A+B$ will be written $(+\: A\: B)$) as any computation
can be written as a tree and filtering solutions is therefore nothing
more than tree reduction. Here are the rules that were used to reduce
the tree:
\begin{itemize}
\item if a number appears in the initial set, it must be used rather
  than built. For example, if we have $\{2,3,5,100\}$, finding 500
  must be done by $(*\; 5\; 100)$ and not $(*\; (+\; 2\; 3)\; 100)$
\item $(+\; A\; B)$ and $(+\; B\; A)$ are identical.
\item $(*\; A\; B)$ and $(*\; B\; A)$ are identical. These two rules have to
  be checked recursively.
\item $(*\; A\; 1)$ and $(/\; A\; 1)$ are $A$
\item $(+\; A\; 0)$ and $(-\; A\; 0)$ are $A$
\item A general reduction rule must be applied to all subtrees that
  contain only $+$ and $-$ operations to put them in a ``canonical''
  form. For example $(+\; (+\; 1\; 4)\; (-\; 6\; (+\; 3\; 2)))$ must
  be reduced to $6$. The 
  algorithm collects all ``positive'' numbers in one list and all ``negative''
  numbers in another list, then suppress all equal numbers or all
  combinations of numbers equal in both lists, and then constructs
  a canonical tree by keeping
  always the smallest positive results in the computation.
\item The same rule applies to subtrees with only $*$ and
  $/$. 
\end{itemize}
These rules have to be applied until the tree is stable.

\begin{figure}[ht!]
  \begin{center}
    \includegraphics[width=7.5cm]{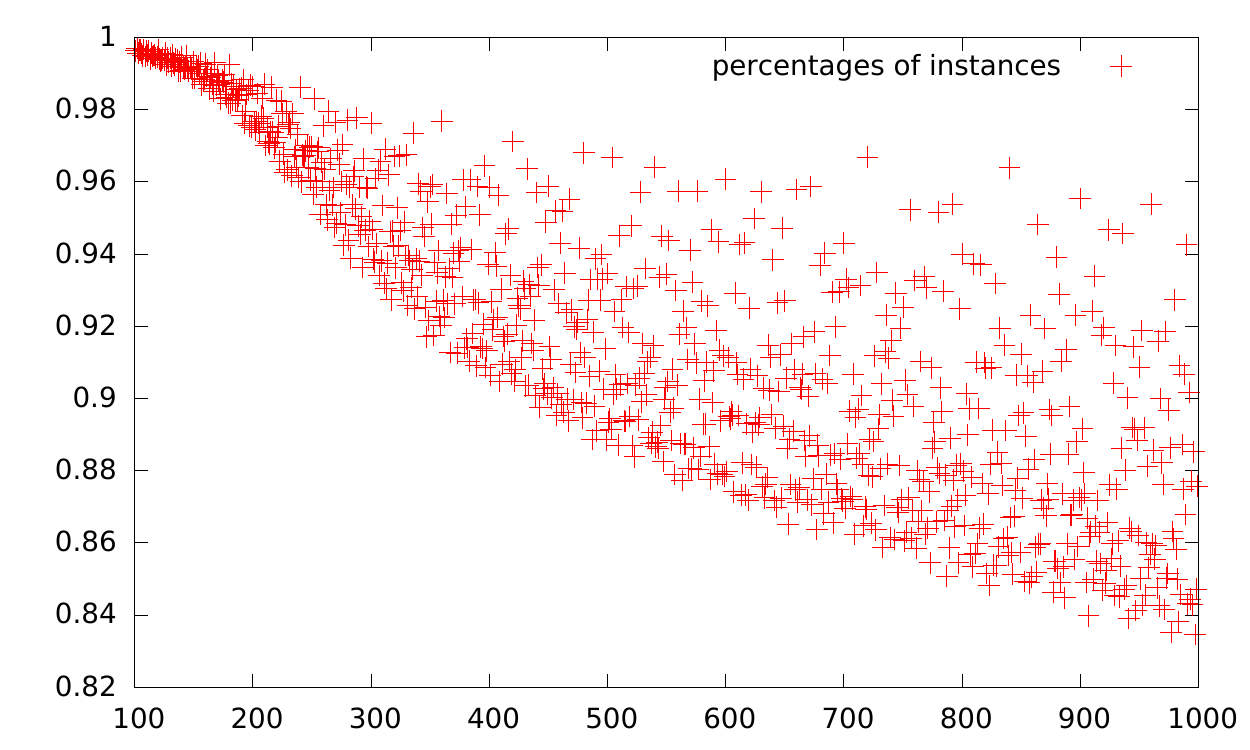}
    \caption{Number of solutions \label{fignbsols}}
  \end{center}
\end{figure}

The results are presented in
figure~\ref{fignbsols}. 
833814 problems (7\%) have 1 solution, 800633 (6.72\%) have 2
solutions, etc\ldots The largest number of different solutions is 232
when the set is $\{2,4,5,6,10,50\}$ and the number to find is 120.

\section{Conclusion}
Tu turn the problem into a challenging one for a human being, this
article proposes different solutions which are easy to use. As the
game has been completely solved for $n=6$, both with the standard set
of numbers and with the extended set of all numbers from 1 to 100, it
is easy to pick numbers and target such that the problem is
challenging for a human being, either by choosing problems which
require a minimal number of operations, or unsolvable problems with
the best findable number at some distance of the target, or using sets
having only prime numbers or large and ``unfriendly'' numbers. Another
solution would be to use more than 6 numbers, and to use a target in a
range above 1000, but it is probably not necessary.
The last solution is to change a little bit the game by adding the
{\em square} operation. We have proved that it is possible with only
6 numbers to find the exact solution for 99.9705\% of the problems
with the target in the 1001--9999 range. This is however much more
difficult for a human being, because using the target is higher, and
the square is not a natural operation to use.

It is more difficult
to turn the game into a challenging problem for a computer. While the
classical depth-first algorithm fails to find a solution in the
allotted amount of time for $n>7$, our algorithm solves the problem
with up to 9 numbers in the set. The $n=10$ problem is out of reach
for an ordinary computer. It
would however be interesting to start a challenge between computers
for $n=10$, or $n=11$ to see what heuristics methods are the best for
solving this problem.
Using the square operation change fundamentally the problem from a
theoretical point of view, because the game might be undecidable.
Proving the undecidability remains an open
challenge, and finding solutions for the currently unsolved problems
(49 instances for the standard set of numbers and standard target)
is still open.
\newpage
\bibliography{compte}

\begin{thebibliography}{17}
\providecommand{\natexlab}[1]{#1}
\providecommand{\url}[1]{\texttt{#1}}
\expandafter\ifx\csname urlstyle\endcsname\relax
  \providecommand{\doi}[1]{doi: #1}\else
  \providecommand{\doi}{doi: \begingroup \urlstyle{rm}\Url}\fi

\bibitem[Alliot(1986)]{Alliot86}
Jean-Marc Alliot.
\newblock Une r\'esolution exhaustive du "compte est bon".
\newblock Communication au groupe des utilisateurs francais de l'Amiga, 1986.

\bibitem[board(1997)]{MPI}
MPI board.
\newblock Mpi-2, 1997.
\newblock URL
  \url{http://www.mcs.anl.gov/research/projects/mpi/mpi-standard/mpi-report-2.0/mpi2-report.htm}.

\bibitem[Buisson(1980)]{Buisson}
Jean-Christophe Buisson.
\newblock A moi compte, deux mots!
\newblock \emph{L'ordinateur individuel}, 20, 1980.

\bibitem[Defays(1990)]{Defays90}
Daniel Defays.
\newblock Numbo: A study in cognition and recognition.
\newblock \emph{The Journal for the Integrated Study of Artificial
  Intelligence, Cognitive Science and Applied Epistemology}, 7\penalty0
  (2):\penalty0 217--243, 1990.

\bibitem[Defays(1995{\natexlab{a}})]{Defays95}
Daniel Defays.
\newblock Numbo: A study in cognition and recognition.
\newblock In Douglas Hofstadter, editor, \emph{Fluid concepts and creative
  analogies: computer models of the fundamental mechanisms of thought}.
  BasicBooks, 1995{\natexlab{a}}.

\bibitem[Defays(1995{\natexlab{b}})]{Defays95f}
Daniel Defays.
\newblock \emph{L'esprit en friche: les foisonnements de l'Intelligence
  Artificielle}.
\newblock Pierre Mardaga, 1995{\natexlab{b}}.
\newblock ISBN: 2-87009-326-8.

\bibitem[Fouquet(2010)]{Fouquet10}
Patrice Fouquet.
\newblock Le compte est bon, March 2010.
\newblock URL \url{http://patquoi.free.fr/lcpdb/}.

\bibitem[Froissart(1984)]{Froissart84}
Marc Froissart.
\newblock Le compte est bon.
\newblock \emph{SVM}, 2:\penalty0 58, 1984.

\bibitem[INRIA(2004)]{Ocaml}
INRIA.
\newblock Ocaml, 2004.
\newblock URL \url{http://caml.inria.fr/ocaml/index.en.html}.

\bibitem[Lemoine and Viennot(2012)]{Kitsune12}
Julien Lemoine and Simon Viennot.
\newblock Kitsune, 2012.
\newblock URL \url{http://kitsune.tuxfamily.org/wiki/doku.php}.

\bibitem[Levinthal(2009)]{Levinthal09}
David Levinthal.
\newblock Performance analysis guide for intel core i7 processor and intel xeon
  5500 processors.
\newblock Intel report, INTEL Corporation, 2009.

\bibitem[Mochel(2003)]{Mochel03}
Jacky Mochel.
\newblock Le compte est bon, April 2003.
\newblock URL \url{http://j.mochel.free.fr/comptebon.php}.

\bibitem[Pin(1998)]{Pin98}
Jean-Eric Pin.
\newblock Le compte est bon.
\newblock Sujets de projets 97-98 de tronc commun informatique de l'Ecole
  Polytechnique de Paris, June 1998.
\newblock Institution: Laboratoire d'Informatique Algorithmique: Fondements et
  Applications.

\bibitem[Tunstall-Pedoe(2013)]{Crosswords}
Williams Tunstall-Pedoe.
\newblock Number games solver faq, 2013.
\newblock URL \url{http://www.crosswordtools.com/numbers-game/faq.php#stats}.

\bibitem[Virtue(2014)]{Virtue14}
G.~Virtue.
\newblock Countdown is 70: Three cheers for the nation's favourite comfort
  blanket.
\newblock \emph{The Guardian}, January 7th 2014.

\bibitem[Wikipedia(2015)]{Countdown}
Wikipedia.
\newblock Countdown (game show), 2015.
\newblock URL \url{http://en.wikipedia.org/wiki/Countdown_(game_show)}.

\bibitem[Zobrist(1970)]{Zobrist70}
Albert~L. Zobrist.
\newblock A new hashing method with application for game playing.
\newblock Technical report~88, University of Wisconsin, Computer Science
  Department, April 1970.

\end{thebibliography}
\end{document}